%% file: main.tex
\documentclass[letterpaper, 10 pt, journal, twoside]{IEEEtran}  % Comment this line out if you need a4paper
\usepackage{comment}
\usepackage{optidef}
\usepackage{dsfont}
\usepackage{booktabs}
\usepackage{multirow}
\usepackage{siunitx}
\usepackage{makecell} % <--- Add this package
\usepackage{acronym}
\usepackage{ral}

\acrodef{MPC}{Model Predictive Control}
\acrodef{MPPI}{Model Predictive Path Integral}
\acrodef{CEM}{Cross Entropy Method}
\acrodef{UAV}{Unmanned Areal Vehicle}
\acrodef{PA-MPPI}{Perception-Aware MPPI}

\newcommand{\R}{\mathbb{R}}
\newcommand{\Hq}{\mathbb{H}}

\newcommand{\reb}[2][\relax]{%
  \ifx#1\relax
    \textcolor{blue}{#2}%
  \else
    \textcolor{blue}{[#1]: #2}%
  \fi
}
\renewcommand{\reb}[2][]{#2}

\title{PA-MPPI: Perception-Aware Model Predictive Path Integral Control for Quadrotor Navigation in Unknown Environments}
\author{Yifan Zhai, Rudolf Reiter, and Davide Scaramuzza% <-this % stops a space
\thanks{Manuscript received: September 5, 2025; Revised December 6, 2025; Accepted January 25, 2026. This paper was recommended for publication by
Editor Soon-Jo Chung upon evaluation of the Associate Editor and Reviewers comments. }% <-this % stops a space
\thanks{
This work was supported by the European Union’s Horizon Europe Research and Innovation Programme under grant agreement No. 101120732 (AUTOASSESS) and the European Research Council (ERC) under grant agreement No. 864042 (AGILEFLIGHT).
These authors are with the Robotics and Perception Group, Department of Informatics, University of Zurich,
Switzerland (\protect\url{https://rpg.ifi.uzh.ch}).
        {Contact: \tt\small dzhai@ifi.uzh.ch}}%
\thanks{Digital Object Identifier (DOI): see top of this page.}
}

\begin{document}

\maketitle

\vspace{-0.25cm}

\input{sections/abstract.tex}

\acresetall % one typically puts this after the abstract and before the conclusion, because some people only read those

% \begin{IEEEkeywords}
% Article submission, IEEE, IEEEtran, journal, \LaTeX, paper, template, typesetting.
% \end{IEEEkeywords}
\vspace{-0.25cm}
\input{sections/supplementary}
\input{sections/intro.tex}
\input{sections/related_work.tex}

\input{sections/preliminaries.tex} % Maybe put before related work
\input{sections/method.tex}

\input{sections/experiments}

\acresetall % one typically puts this after the abstract and before the conclusion, because some people only read those

\input{sections/conclusion.tex}

\newpage
\bibliographystyle{IEEEtran}
\bibliography{reference}

\vfill

\end{document}

%% file: sections/abstract.tex
\begin{abstract}
%%%%%%%%%%%%%%%%%%%%%%%%%%%%%%%%%%%%%%%%%%%%%%
% 1. Motivation: Task description / why is it important?
Quadrotor navigation in unknown environments is critical for practical missions such as search-and-rescue. 
% 2. Challenge: Why is problem so difficult?
Solving it requires addressing three key challenges: path-planning in non-convex free space due to obstacles, satisfying quadrotor-specific dynamics and objectives, and exploring unknown regions to expand the map.
% 3. Trends: How does SotA approach it? What's missing?
Recently, the Model Predictive Path Integral (MPPI) method has emerged as a promising solution that solves the first two challenges. By leveraging sampling-based optimization, it can effectively handle non-convex free space while directly optimizing over the full quadrotor dynamics, enabling the inclusion of quadrotor-specific costs such as energy consumption. However, MPPI has been limited to tracking control that only optimizes trajectories in a small neighbourhood around a reference trajectory, as it lacks the ability to explore unknown regions and plan alternative paths when blocked by large obstacles.
% 4. Method: How do you solve it? Contributions!
To solve this issue, we introduce Perception-Aware MPPI (PA-MPPI). Here, perception-awareness is characterized by planning and adapting the trajectory online based on perception objectives. Specifically, when the goal is occluded, PA-MPPI's perception cost biases trajectories that can perceive unknown regions. This expands the mapped traversable space and increases the likelihood of finding alternative paths to the goal.
% 5. What do the results say?
Through hardware experiments, we demonstrate that PA-MPPI, running at \SI{50}{\hertz}, performs on par with the SOTA quadrotor navigation planner for unknown environments in our challenging test scenarios. 
In addition, we demonstrate that PA-MPPI can be used as a safe and robust action policy for navigation foundation models, which often provide goal poses that are not directly reachable.
%%%%%%%%%%%%%%%%%%%%%%%%%%%%%%%%%%%%%%%%%%%%%%

\end{abstract}

%% file: sections/supplementary.tex
\hypersetup{
  colorlinks=true,
  urlcolor=ethblue
}

\section*{Supplementary Materials}

\noindent
\faVideo\hspace{0.6em}\text{Video:}  
\url{https://youtu.be/DNUK2gUcVIM}

%% file: sections/intro.tex
\begin{comment}
Intro structure:
1. What is the problem?
2. Why is it important?
3. Why is the problem hard? What makes it challenging?
4. How far has existing work come? What is the next frontier?
5. Why hasn’t the problem been solved? What is the stumbling block?
6. What does our paper contribute?
7. What is the key idea? What is the magic trick? What is the new insight or technique that enables us to advance the frontier? Basically, what is that one thing that would make the reader result in a "Aha! that's the idea!"
8. What do the experiments say?
\end{comment}

\section{Introduction}

\begin{figure}[!t]
\centering
\includegraphics[width=\columnwidth]{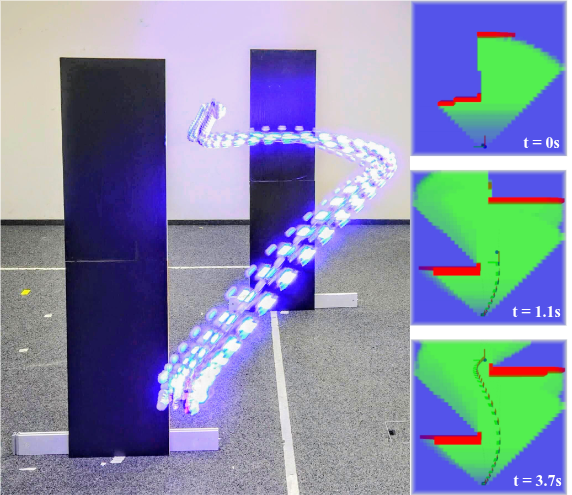}
\caption{The \ac{PA-MPPI} controller navigating to the goal pose while avoiding obstacles in a previously unknown environment. The controller simultaneously controls the quadrotor at \SI{50}{\hertz} and optimizes the perception objective based on an online-updated 3D map constructed from onboard observations.}
\vspace*{-0.2cm}
\label{fig1}
\end{figure}

% 1) What is the problem + why is it important?
\IEEEPARstart{E}{nabling} quadrotors to navigate to a goal in unknown environments autonomously is critical for practical missions such as search-and-rescue, infrastructure inspection, and exploration~\cite{recalde2022system,xing2023autonomous,papachristos2019localization}. It is also relevant for the safe deployment of navigation foundation models, which provide navigation waypoints or goals in previously unseen environments~\cite{sridhar2023nomad, cheng2024navila}.
% 2) why is it hard, what makes it challenging:
Achieving this capability requires addressing three key challenges: (i)~non-convex constraints: cluttered environments create non-convex free space, which complicates gradient-based optimization; (ii)~quadrotor-specific dynamics and costs: planned trajectories must satisfy dynamic feasibility while optimizing costs such as effort and energy; and (iii)~mapping in unknown environments: since the environment must be mapped online using onboard perception, successful navigation must incorporate exploration and mapping of unknown regions to find a feasible path to the goal.

% 3) How far has existing work come?
A common approach is a hierarchical planner–controller architecture~\cite{cieslewski2017rapid,naazare2019application,liu2024integrated,achtelik2013path, super}, where a global planner computes trajectories subsequently tracked by a local controller. \reb[R5.3]{However, this separation often results in conservative, or dynamically infeasible plans, as quadrotor models and related costs and constraints need to be approximated~\cite{hehn2015realtime}}.

% 4) What is the next frontier?
Model Predictive Path Integral (MPPI) control has recently emerged as a promising alternative. By employing sampling-based optimization, MPPI can navigate non-convex and nonsmooth free space in the presence of obstacles while directly optimizing control inputs on quadrotor-specific dynamics and costs~\cite{saska2024MPPI, mohamed2020mppi}.
% 5) Why hasn't the problem been solved? What is the stumbling block?
\reb[R5.3]{However, MPPI has been largely confined to following reference trajectories, demonstrating only limited local optimization near the reference, as it lacks both global map awareness and the capability to explore unknown areas and plan successful trajectories.}

% 6) What does our paper contribute?
In this work, we introduce the Perception-Aware Model Predictive Path Integral (PA-MPPI) controller, tightly integrated with a perception and mapping module, to afford standard MPPI with the capability of navigating unknown environments without external references.
% 7) What is the key idea? What is the magic trick? What is the new insight or technique that enables us to advance the frontier? Basically, what is that one thing that would make the reader result in a "Aha! that's the idea!"
This is achieved by extending standard MPPI with perception awareness, characterized by adapting control inputs and trajectory optimization based on both current and future perception of the environment \cite{xing2023autonomous,falanga2018pampc,Sarvaiya2024hpampc}. 
%Prior works have demonstrated the effectiveness of perception-aware costs for maintaining visual targets within the onboard camera’s field of view \cite{xing2023autonomous,falanga2018pampc,Sarvaiya2024hpampc}. PA-MPPI expands on this by introducing an additional perception cost. 
Using the current map of the environment, PA-MPPI introduces a novel perception cost that evaluates sampled trajectories based on their potential to perceive unknown regions in the goal direction, thereby guiding the optimized trajectory to map unknown regions that lead to the goal.

% 8) What does the experiment say:
\reb[R5.1, R7.1]{We validate PA-MPPI in simulated and real-world experiments, demonstrating that perception-awareness enables MPPI to navigate through complex, unknown environments with performance comparable to SUPER~\cite{super}, the current state-of-the-art planner for safety-assured quadrotor navigation in unknown environments.}

Our contributions are threefold:
\begin{itemize}
    \item \textbf{Novel Perception-Aware MPPI Formulation:} We propose a cost function that exploits the current environment map to guide trajectory optimization toward frontiers that help perceive unknown regions towards the goal, allowing MPPI to navigate without external references.
    \item \textbf{Integrated Framework:} We present an integrated framework that tightly couples sensing, mapping, and a high-performance MPPI implementation, running at \SI{50}{\hertz} for real-time quadrotor control.
    \item \textbf{Hardware-in-the-Loop Validation:} \reb[R7.1, R5.1]{Our experiments show that \ac{PA-MPPI} performs comparably to the current SOTA planner, SUPER \cite{super}.} We further demonstrate that using \ac{PA-MPPI} as the action policy enables robust deployment of navigation foundation models.
\end{itemize}

%%%%%%%%%%%%%%%%%%%%%%%%%%%%%%%%%%%%%%%%%%%%%%%%%%%%%%%%%%%%%%%

%% file: sections/related_work.tex
\section{Related Work}
\textbf{Obstacle Avoidance in Cluttered Environments.}
In classical sense-think-act architectures, a high-level planner is responsible for more complex objectives and generates a safe trajectory, which is then passed to a fast lower-level controller~\cite{mellinger2011,hehn2015realtime}. Often, planners operate in a discrete or low-dimensional control/state space, for example, using graph search~\cite{cieslewski2017rapid,naazare2019application,liu2024integrated} or rapidly exploring random trees~\cite{achtelik2013path}. \reb[R5.3]{However, with a reduced state or control space, trajectories are suboptimal and conservative, and the hierarchical decomposition introduces an interaction between the controller and the planner. In particular, the novel perception-awareness requires a cost associated with orientation that cannot be accounted for in classical higher-level planners, such as those in \cite{mellinger2011,hehn2015realtime}.}

Optimization-based approaches often simultaneously plan safe trajectories and control the actuators. However, optimization-based motion planning in the presence of obstacles is challenging due to the induced nonconvexity and nonsmoothness~\cite{zhou2021raptor,yu2022cpa,marcucci2023motion,gao2024realtime,reiter2024progressive, EGO-Planner}. 
%A recent approach~\cite{marcucci2023motion} and, similarly~\cite{yu2022cpa}, exploit graphs of convex sets to cast collision-free planning as a tight convex program, scaling to high-dimensional environments~\cite{marcucci2023motion}. 
The authors in~\cite{yu2022cpa,gao2024realtime,reiter2024progressive} propose efficient algorithms for ellipsoidal or cubic obstacles, and \cite{recalde2022system} shows a real-world implementation for \acp{UAV}.
However, such derivative-based formulations typically rely on smooth functions, which can be difficult to obtain directly from raw depth observations in complex, highly non-smooth environments. \reb[R5.4]{To circumvent this, \cite{yadav2023receding_horizon} generates a number of candidate reference trajectories, and assigns cost based on collision and proximity to the goal. However, the optimization that generates the candidates does not consider actuation limits, which requires additional checks and potentially multiple iterations of relaxation and re-optimization.}

Beyond optimization-based approaches, learned depth-camera-based policies have shown sufficient navigation performance when trained end-to-end from visual inputs. The authors in~\cite{loquercio2021learning} demonstrated high-speed flight in real-world environments using policies trained in simulation. More recently, vision transformer-based policies have been explored for end-to-end quadrotor obstacle avoidance using onboard computation~\cite{bhattacharya2025vision}. While such approaches improve the closed-loop cost disregarding collisions, they often lack the safety-relevant explicit constraint handling and adaptability to new environments provided by optimization-based methods.

\textbf{Perception-Aware Navigation.}
A growing number of works integrates perception objectives directly into planning and control, either to acquire task-relevant information or to maintain estimation quality. Perception-aware \ac{MPC} aligns the robot’s viewpoint with features to maximize visibility during challenging maneuvers~\cite{falanga2018pampc,lee2020aggressive,xing2023autonomous}. Information-theoretic surrogates such as Fisher Information Fields provide differentiable maps for actively choosing informative viewpoints~\cite{zhang2020fisher,lim2023fisher}. In exploration and inspection, planners explicitly reason about localization uncertainty to select safe, observable trajectories~\cite{papachristos2019localization}.

The research direction of our work treat unobserved space as potentially blocked or even hazardous within partial maps, thus requiring its goal-targeted exploration~\cite{cieslewski2017rapid,zhou2021raptor,yu2022cpa}. The high-level planners in~\cite{zhou2021raptor,yu2022cpa} enforce perception awareness along an optimistic trajectory by monitoring potentially hazardous regions, while~\cite{cieslewski2017rapid} plans to explore the frontier of unknown space via a Dijkstra graph-search. The current state-of-the-art planner, SUPER \cite{super}, simultaneous plans a exploratory trajectory, guided by A* graph-search in both free and unknown space for exploration and high speed flight, and a backup trajectory in known free space for emergency stops. As these ideas typically resort to a separate planner, they motivate our use of perception-driven costs in a single controller that rewards the line of sight to the goal and prevents motion into unknown regions, with a focus on highly cluttered environments obtained from depth images.

\textbf{Algorithms for Model Predictive Control.}
\reb[R5.2]{In the following, the proposed perception-aware \ac{MPPI} algorithm is motivated, which extends classical MPPI~\cite{theodorou2012relative} by a perception cost.} 

\paragraph{Derivative-based MPC.}
When objectives and constraints admit smooth approximations, derivative-based MPC offers strong local convergence and tight constraint handling. Progressive smoothing and continuation strategies have been proposed to navigate nonconvex obstacle costs~\cite{reiter2024progressive}, conceptually paralleling equal to annealing in sampling-based MPC~\cite{xue2024dialmpc}. 
A recent approach proposed an algorithm with an external active set solver for cluttered point-cloud obstacles~\cite{gao2025semiinfinite}, which achieves a feasible average but has an ample worst-case computation time.
Still, in settings where, in addition to obstacles represented by raw depth maps, the cost functions are also highly nonsmooth, constructing reliable differentiable surrogates remains challenging. Sampling methods showed superior performance in these settings~\cite{suh2022differentiable}.

\paragraph{Sampling-based MPC.}
\reb[R5.2]{Sampling-based MPC methods, such as MPPI, optimize control sequences by Monte Carlo rollouts instead of using local gradients,} making them attractive for nonconvex, nonsmooth objectives and dynamics. The most straightforward but largely inefficient random shooting method purely randomizes actions~\cite{piovesan2009randomized}. The \ac{CEM}~\cite{rubinstein1997optimization} iteratively refines a distribution toward high-performing controls via an elite-only update and has been widely adopted as a trajectory optimizer and within model-based RL pipelines~\cite{chua2018deep}. \Ac{MPPI} control~\cite{kappen2005linear,theodorou2012relative,theodorou2015nonlinear} uses all samples via weighting to iteratively refine an trajectory~\cite{Williams2017MPPI}. \reb[R5.2]{\Ac{MPPI} has recently been pushed to demanding robot platforms, including agile UAVs~\cite{saska2024MPPI} and whole-body locomotion~\cite{alvarez2025realtime, xue2024dialmpc}. However, these works rely on external references, such as a reference quadrotor trajectory or quadruped walking gait joint position reference. PA-MPPI, on the other hand, enables reference-free navigation via the novel perception-aware cost.}

%% file: sections/preliminaries.tex
\section{Preliminaries}
% In the following, the notation of this paper is introduced in Sect.~\ref{sec:notation}, followed by a brief description of the \ac{MPC} framework in Sect.~\ref{sec:mpc}.

\subsection{Notation}
\label{sec:notation}
We introduce two reference frames: \( W \), the fixed world frame, whose \( z \)-axis is gravity-aligned, and \( B \), the quadrotor body frame, whose \( x \)-axis aligns with the onboard camera's principal axis. In this paper, vectors and matrices are written in bold, with matrices indicated by capital letters. Each vector carries a subscript specifying the frame in which it is expressed and its endpoint. For instance, \reb[R7.5]{\(\bm{p}_{WB}\in\R^3\)} denotes the position of the body frame \( B \) relative to the world frame \( W \), and \(\bm{R}_{WB}\) denotes the rotation from frame \( B \) to \( W \). We use $\bm{y}_{1:N}\in\mathbb{R}^{N n_y}$ for the vectorization of vectors~$\bm{y}_1,\ldots, \bm{y}_N$, with $\bm{y}_i\in\mathbb{R}^{n_y}$. The quaternion algebra is~$\Hq$ and $q\in\Hq_1:=\left\{q\in\Hq\mid\vert \|q\|=1\right\}$.

\subsection{Model Predictive Control}
\label{sec:mpc}
Given an environment, often formulated as a Markov decision process (MDP), with the states~$\bm{x}\in\R^{n_x}$, controls~$\bm{u}\in\R^{n_u}$, a stage cost~$l:\R^{n_x}\times\R^{n_u}\rightarrow\R\cup\{\infty\}$, a discount factor~$\gamma$, and a stochastic model
$P: \R^{n_x} \times \R^{n_u} \rightarrow \mathrm{Dist}(\R^{n_x})$,
\ac{MPC} provides a means to yield nearly optimal (sufficiently suboptimal) controls by solving an approximated implicit version of the MDP online, local at the current state~$\hat{\bm{x}}\in\R^{n_x}$, cf.~\cite{reiter2025synthesis} for details. \ac{MPC} typically uses a simplified deterministic model~$f:\R^{n_x} \times \R^{n_u}\rightarrow\R^{n_x}$, possibly simplified stage cost~$\ell:\R^{n_x}\times\R^{n_u} \rightarrow\mathbb{R}$ and an approximation~$\bar{V}:\R^{n_x}\rightarrow\mathbb{R}$ of the optimal value function~$V:\R^{n_x} \rightarrow\mathbb{R}$. Moreover, an open-loop trajectory~$\bm{u}_{0:H-1}$ with open-loop states~$\bm{x}_i\in\R^{n_x}$, actions~$\bm{u}_i\in\R^{n_u}$ and horizon~$H$ is optimized, instead of a policy. The resulting single shooting MPC optimization problem is therefore
\begin{align}
\label{eq:mpc}
\begin{split}
    &\min_{\bm{u}} \;\bar{V}(\bm{x}_{H};\bm{p})+\sum_{k=0}^{H-1} \ell(\bm{x}_k, \bm{u}_k;\bm{p}) \\
    & \text{with}\quad \bm{x}_0=s,\;\bm{x}_{k+1}= f(\bm{x}_k, \bm{u}_k),\;0\leq k < H,
\end{split}
\end{align}
where constraints are approximated in the value and cost function. We denote the dependency of the cost function on external parameters, such as obstacle parameters or the goal by~$\bm{p}\in\R^{n_p}$.
Unlike gradient-based algorithms for solving the \ac{MPC} problem~\eqref{eq:mpc}, which typically rely on first or second-order derivatives, \ac{MPPI} uses Monte Carlo sampling-based optimization. 
Despite the initial derivation of \Ac{MPPI} control for the stochastic system~\cite{kappen2005linear,theodorou2012relative,theodorou2015nonlinear}, it is often used for the deterministic counterpart~\eqref{eq:mpc}, where noise is added as part of the optimization algorithm~\cite{homburger2025optimality}.

%% file: sections/method.tex
\section{Methodology}
\begin{figure}[!t]
\centering
\includegraphics[width=\columnwidth, , trim=0 0 0 0, clip]{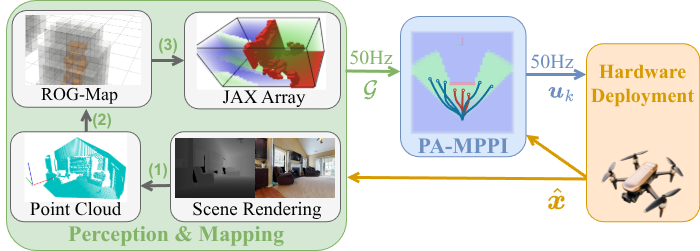}
\caption{Illustration of the control stack. The perception and mapping module has three parts: (1) a point cloud is generated from the rendered depth image and transformed to the world frame, (2) the ROG-Map \cite{ROG-Map} updates the occupancy map with the new point cloud, and (3) the ROG-Map is converted to a JAX array as input to PA-MPPI.}
\vspace*{-0.4cm}
\label{fig:stack}
\end{figure}

 A graphical overview of the integrated control stack is shown in Fig. \ref{fig:stack}. As illustrated, PA-MPPI receives an occupancy grid from the perception module and optimizes a trajectory as control sequences, which are directly executed by the quadrotor. 
 %In the hardware-in-the-loop setting, the quadrotor state is used to render depth images, which are then processed by the perception pipeline to update the occupancy grid in real time. 
 The proposed PA-MPPI algorithm utilizes a quadrotor dynamics model, as described in Sect.~\ref{sec:dynamics}.  A detailed description of the \ac{MPPI} formulation is given in Sect.~\ref{sec:mppi}. The perception and mapping part, and the cost definition are detailed in Sect.~\ref{sec:mapping} and~\ref{sec:cost}, respectively.

\vspace{-0.3cm}
\subsection{Quadrotor Dynamics}
\label{sec:dynamics}

\reb[R7.5]{In addition to the quadrotor's position \(\bm{p}_{WB}\),  we define orientation, and linear velocity in the world frame by \(\bm{q}_{WB}\in\Hq_1\), and \(\bm{v}_{WB}\in\R^3\), respectively,} and the quadrotor's angular velocity in the body frame by \(\bm{\omega}_{B}\in\R^3\).  
The collective thrust and the corresponding thrust vector in the body frame are defined as
$c=c_1+\ldots+c_{N_\mathrm{rot}}$ and 
$ \bm{c}_B = \begin{bmatrix} 0 & 0 & c \end{bmatrix}^\top$,
where \(c_i\) is the thrust generated by the \(i\)-th of $N_\mathrm{rot}$ motors.
The quadrotor mass is~\(m\) and \(\bm{g}_W\) is the gravity vector in the world frame.  
Finally, the diagonal moment of inertia matrix is \(\bm{J}\in\R^{3\times3}\), and the body torque is~\(\bm{\tau}_B \in \R^3\).  
The quadrotor dynamics can then be expressed as
\input{equations/dynamics}

The lowest-level flight controller tracks the zero-order hold PA-MPPI control $\bm{u}_{t}=\begin{bmatrix} c_t & \bm{\omega}_{B,t} \end{bmatrix} ^\top$. 
To ensure a feasible total thrust and body rate at each timestep, we follow the single motor thrust clipping in \cite{saska2024MPPI}, using the motor thrust limits to acquire clipped control input $u_t^{\text{clip}}$, which is then used by PA-MPPI to simulate the dynamics~\eqref{eq:dynamics} via forward Euler integration. 
\vspace{-0.4cm}
\subsection{MPPI formulation}
\label{sec:mppi}
In the MPPI framework, at each timestep~$k$, $N$ parallel trajectories of $H$ steps are sampled by adding multivariate Gaussian noise to the nominal control input~$\bm{u}^{\text{nom}}_{k:k+H}$. The perturbed control sequences $\bm{u}_{k:k+H}^j$, $j=1, ..., N$, are then rolled out from the quadrotor state $\bm{x}_k$ using the model~\eqref{eq:dynamics}. Each sampled trajectory $\bm{x}_{k:k+H}^{j}$ is then evaluated by summing the per-step costs
$
\mathcal{L}^j = \sum_{i=k}^{k+H-1} \ell(\bm{x}_i^j, \bm{u}_i^j) + \bar{V}(\bm{x}^j_H).
$
The optimized control sequence is calculated using the exponentially weighted average based on the summed cost
\reb[R7.6]{
\[
\bm{u}_{k:k+H-1} = \sum_{j=1}^{N} w^j\bm{u}_{k:k+H-1}^j,\;
w^j = \frac{\exp{(-\frac{\mathcal{L}^j - \mathcal{L}^{\text{min}}}{\lambda})}}{\sum_{j=1}^{N} \exp{(-\frac{\mathcal{L}^j - \mathcal{L}^{\text{min}}}{\lambda})}},
\]}
where $\mathcal{L}^{\text{min}}$ is the lowest summed cost out of the $N$ rollouts, and $\lambda$ is the temperature parameter. A low $\lambda$ assigns higher weight to the best-performing rollout, while higher values assign more uniform weights to all rollouts \cite{Williams2017MPPI}. The first action of the averaged sequence is then executed, and the optimization process repeats in the next time step.

Recent works \cite{saska2024MPPI,xue2024dialmpc,howell2022} have shown success in decoupling the prediction time step size~$\Delta t_{\text{pred}}$, and the control step size~$\Delta t_{\text{ctrl}}$. With~$\Delta t_{\text{pred}} > \Delta t_{\text{ctrl}}$, the policy rollouts can predict over a longer real-time horizon for the same number of forward simulation steps, allowing optimization of actions further into the future. Since the optimization loop executes the first action at control frequency and the actions in the sequence are spaced by $\Delta t_{\text{pred}}$, to reuse the remaining actions as $\bm{u}^{\text{nom}}$ for the next iteration, the control sequence is linearly interpolated and shifted by $\Delta t_{\text{ctrl}}$, then down-sampled at $\Delta t_{\text{pred}}$.

While the terminal value~$\bar{V}$ function should in principle resemble the optimal value function and account for a recursive feasible safe set, cf.~\cite{reiter2025synthesis}, we only consider a simple terminal hovering safe set with zero velocity and resort to a long enough planning horizon to diminish its influence on the open loop cost.
\vspace{-0.2cm}
% First, the desired body torque to track the input is calculated via
% \[
% \bm{\tau}^d_B = \bm{J} \cdot \dot{\bm{\omega}}_B^d + \bm{\omega}_B \times \bm{J} \cdot \dot{\bm{\omega}}_B,
% \]
% where \(\dot{\bm{\omega}}_B^d\) is the desired angular acceleration in the body frame, approximated using the timestep~\(\Delta t\) by 
% \[
% \dot{\bm{\omega}}_d = \frac{1}{\Delta t} (\bm{\omega}_d - \bm{\omega}).
% \]
% Using the control allocation matrix~\(\bm{B}\), the desired single rotor thrusts~\(\bm{t}^d\) can be computed via
% \[
% \bm{t}^d = \begin{bmatrix} c_1 & c_2 & c_3 & c_4 \end{bmatrix}^\top = \bm{B}^{-1} \begin{bmatrix} c^d \\ \bm{\tau}^d_B \end{bmatrix}.
% \]
% Based on the motor thrust constraints~\(t_{\text{min}}\) and~\(t_{\text{max}}\), the desired single motor thrusts are clipped with
% \[
% \bm{t}^{\text{clip}} = \text{clip}(t_{\text{min}}, \bm{t}^d, t_{\text{max}}).
% \]
% The clipped collective thrust and body torque are thus
% \[
% \begin{bmatrix} c^{\text{clip}} & \bm{\tau}_B^{\text{clip}}\end{bmatrix} = \bm{B} \cdot \bm{t}^{\text{clip}}
% \]
% which is used by PA-MPPI to simulate the dynamics~\eqref{eq:dynamics} via forward Euler integration.

\subsection{Perception \& Environment Mapping}
\label{sec:mapping}
% \begin{figure*}[!t]
%     \centering
%     \subfloat[Reconstructed Scene]{%
%         \includegraphics[width=0.245\textwidth, trim=125 275 125 225, clip]{figures/mapping/mapping_rgb.png}%
%         \label{fig:sub1}%
%     }\hfill
%     \subfloat[Occupied Voxels]{%
%         \includegraphics[width=0.245\textwidth, trim=125 275 125 225,clip]{figures/mapping/mapping_occupied.png}%
%         \label{fig:sub2}%
%     }\hfill
%     \subfloat[Free Voxels]{%
%         \includegraphics[width=0.245\textwidth, trim=125 275 125 225,clip]{figures/mapping/mapping_free.png}%
%         \label{fig:sub2}%
%     }\hfill
%     \subfloat[Unknown Voxels]{%
%         \includegraphics[width=0.245\textwidth, trim=125 275 125 225,clip]{figures/mapping/mapping_unknown.png}%
%         \label{fig:sub2}%
%     }
%     \caption{Mapping example of a scene from the Habitat Matterport dataset~\cite{ramakrishnan2021hm3d} 
% using four consecutive RGB--D images with the corresponding camera poses. (a) Scene reconstruction obtained by projecting RGB values onto the depth point cloud, with camera poses and the mapping bounding box overlaid. (b)(c)~Occupied and free voxels extracted from the Octomap. (d) Unknown voxels, which naturally appear outside the camera's FoVs or in areas occluded by objects.}
%     \label{fig:mapping}
% \vspace{-0.5cm}
% \end{figure*}

We use a depth sensor on the quadrotor to continuously build a 3D map of the environment during navigation. From each depth image, we reproject a point cloud into the world frame using the camera pose. The point clouds are then inserted into a ROG-Map \cite{ROG-Map}, which efficiently aggregates all past observations into a 3D occupancy grid representation, in which each voxel has one of three states: occupied, free, or unknown, with integer values $\{1, 0, -1\}$ respectively. Formally, the 3D occupancy grid is defined as 
$
\mathcal{G} \in \{-1, 0, 1\}^{X \times Y \times Z}.
$
We denote by \reb[R7.5]{$\mathcal{G}(\bm{p}_{WB})$} the operation that looks up the voxel corresponding to position \reb[R7.5]{$\bm{p}_{WB}$} and returns its value.
 The mapping pipeline is capable of processing depth images and updating the occupancy grid at \SI{50}{\hertz} for a $5 \times 5 \times 2\si{\meter}$ grid with a voxel resolution of \SI{0.1}{\meter}.

\subsection{Optimization Cost Definition}
\label{sec:cost}
\reb[R7.2]{
At each timestep $k$ within the planning horizon $H$, $k=0,...,H-1$, the stage cost of~\eqref{eq:mpc} for the navigation task consists of the following terms:}
\begin{align*}
% \ell_{\text{goal}} &= \big(-c_{\text{pos}} + c_{\psi} \cdot |\Delta {\psi} | \big) \cdot \exp \big(- \| \bm{d}_{\text{goal}} \|^2 \big),\\
\ell_{\text{goal}} & = -c_\text{goal} \cdot \max(0.0, \bm{d}_0 - \bm{d}_k)\\
\ell_{\text{goal}, H-1} &= -c_{\text{goal}, H-1} \cdot \max(0.0, \bm{d}_0 - \bm{d}_{H-1})\\
\ell_{\text{act}} &= \| \bm{u}\|^2_R + \| \Delta\bm{u} \|^2_{R_\Delta},\\
\ell_{\text{vel}} &= \exp(-c_\text{vel} \cdot \bm{d}_k^2) \cdot \|\bm{v}_k\|^2 \\
\ell_{\text{progress}} &= -c_\text{progress} \cdot \|\bm{p}_{WB, i} - \bm{p}_{WB, i+1}\| \\
\ell_{\text{collision}} &= c_\text{collision} \cdot \mathds{1}_{\{\mathcal{G}(\bm{p}_{WB}) \neq 0\}},\\
\ell_{\text{perception}} &= c_{\text{PoI}} \cdot (1-\langle \hat{\bm{\mathrm{x}}}_{WB}, \hat{\bm{d}}_{\text{goal}} \rangle)^2 \cdot \mathds{1}_{\{ \| \bm{d}_{\text{goal}} \| > c_\text{thresh}\}} \\
&\quad + c_\text{occupied} \cdot \mathds{1}_{\{ \mathcal{G}(\bm{r}(t^*))=1\}} \\
&\quad + c_\text{unknown} \cdot \mathds{1}_{\{ \mathcal{G}(\bm{r}(t^*))=-1\}} 
\end{align*}

and the quantities: 
\begin{align*}
\bm{d}_i & = \|\bm{p}_{WB, i} - \bm{p}_{\text{goal}}\|, \quad \hat{\bm{d}}_{\text{goal}} = \frac{\bm{d}_{\text{goal}}}{\|\bm{d}_{\text{goal} \|}}, \\
\hat{\bm{\mathrm{x}}}_{WB} & = \bm{R}_{WB} \cdot \bm{e}_1.
% \Delta{\psi} &= \mathrm{atan2}(\sin(\psi_{WB} - \psi_{\text{goal}}), \cos(\psi_{WB} - \psi_{\text{goal}})).
\end{align*}

The terminal value function is $\bar{V}(x)=c_\text{safe}\mathds{1}_{\{\|\bm{v}_{WB}\|>\underline{\bm{v}}\}\vee \|\bm{\omega}_{B}\|>\underline{\bm{\omega}}},$
with bounds~$\underline{\bm{v}}$ and $\underline{\bm{\omega}}$, and a large penalty~$c_\mathrm{safe}$ to account for safe set (hovering).

\begin{figure}[!t]
    \centering
    \subfloat[Given only one depth observation, there is no position in known free space that has a direct line of sight to the goal. Trajectories receive a reward for exploring unknown regions (blue trajectories) or a penalty for facing obstacles (red trajectories).]{%
        \parbox{0.45\textwidth}{ % <-- controls caption width
            \centering
            \includegraphics[width=0.3\textwidth]{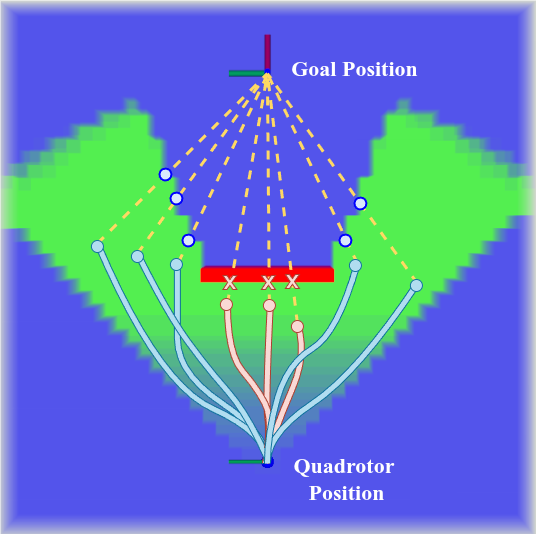}%
        }%
        \label{fig:raytrace_sub1}%
    }\hfill
    \subfloat[\footnotesize After moving to a new position while mapping the environment, there is a direct line of sight from the current position to the goal, and the second-phase cost is active, encouraging the quadrotor to directly reach the goal, and the perception cost is not assigned anymore.]{%
        \parbox{0.45\textwidth}{ % <-- controls caption width
            \centering
            \includegraphics[width=0.3\textwidth, trim=30 0 0 0,clip]{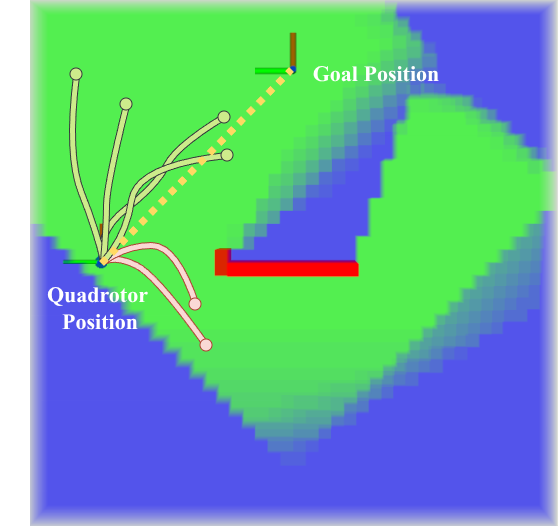}%
        }%
        \label{fig:raytrace_sub2}%
    }
    \caption{\footnotesize A top-down visualization of the ray-tracing in perception cost calculation, showing the occupied voxels (red), free voxels (green), and unknown voxels (blue), and sampled trajectories on which ray-tracing is performed.}
    \label{fig:raytrace}
    \vspace*{-0.6cm}
\end{figure}

\reb[R7.2, R5.4]{The stage costs have two distinct phases: when the goal is not within the direct line of sight of the quadrotor, as shown in Fig. \ref{fig:raytrace_sub1}, to encourage exploration, the cost is 
$\ell = \ell_{\text{goal}}+\ell_{\text{goal}, H}+\ell_{\text{act}}+\ell_{\text{collision}} + \ell_{\text{perception}}$, with $c_\text{goal} = 0.125$ and $c_{\text{goal},H-1}=10$; when the goal is within the direct line of sight, as shown in Fig. \ref{fig:raytrace_sub2}, to quickly reach the goal and hover, the cost is 
$\ell = \ell_{\text{goal}}+\ell_{\text{act}}+\ell_{\text{collision}} + \ell_{\text{progress}} + \ell_{\text{vel}}$, with $c_\text{goal} = 5.0$.
The goal cost~$\ell_{\text{goal}}$ encourages getting closer to the goal compared to the starting position of the current horizon. For the last step of the horizon $k=H-1$, an additional cost $\ell_{\text{goal}, H-1}$ is given to weight the end position of the horizon more, similar to the intermediate point cost in \cite{yadav2023receding_horizon}. This prevents greedy behavior and allows the policy to detour and go around obstacles if that brings the policy closer to the goal in the end.} The action cost~$\ell_{\text{act}}$ penalizes the magnitude and change in control inputs, similar to the implementation in \cite{saska2024MPPI}. The velocity penalty $\ell_{\text{vel}}$ encourages the quadrotor to slow down near the goal, and the progress cost $\ell_{\text{progress}}$ encourages fast movement when a straight line path to the goal is available.

The collision cost~$\ell_{\text{collision}}$ is a binary value weighted by a large constant \(c_\text{collision}\). The indicator function $\mathds{1}_{\{\mathcal{G}(\bm{p}_{WB}) \neq 0\}}$ returns 1 if the quadrotor's current position lies outside the set of free voxels defined in Section \ref{sec:mapping}. This large penalty enforces the quadrotor not only to avoid collisions with known obstacles but also to refrain from entering unknown regions, which is critical for ensuring safety when navigating in unknown environments.

The perception cost, $\ell_{\text{perception}}$, consists of two components. The first, weighted by $c_{\text{PoI}}$, encourages alignment of the quadrotor’s x-axis (coinciding with the depth camera’s principal axis) with the direction of the goal position (point of interest), thereby maximizing the goal’s visibility within the image frame \cite{falanga2018pampc,xing2023autonomous}. As the quadrotor approaches the goal (distance below $c_{\text{thresh}}$), this term becomes inactive. The last two terms in $\ell_{\text{perception}}$ represent two mutually exclusive cases of the quadrotor position with respect to the mapped region of the environment. We define a ray that starts from the quadrotor position $\bm{p}_{WB}$ and ends at the goal position $\bm{p}_{\text{goal}}$ as:
$
\bm{r}(t) = \bm{p}_{WB} + t\cdot \bm{d}_{\text{goal}}, \; 0 \leq t \leq 1.
$
Since $\bm{p}_{WB}$ is constrained to the known free space due to the collision cost, there exists a $t^*$ such that the ray either exits the free space at an obstacle or unknown space, as shown in Fig.~\ref{fig:raytrace_sub1}. If $\bm{r}(t^*)$ lies in an occupied voxel, a cost $c_{\text{occupied}}$ is assigned. If $\bm{r}(t^*)$ falls in an unknown voxel, then it suggests an exploration frontier towards the goal is present, and a negative cost $c_{\text{unknown}}$ is given. When there is a direct line of sight from the current position to the goal, the second-phase cost is active, as there is no need for exploration, and no perception cost is assigned. As illustrated in Fig.~\ref{fig:raytrace}, this ray-tracing term favors sampled trajectories that either explore unknown regions when the goal is blocked by obstacles (Fig.~\ref{fig:raytrace_sub1}) or move directly toward the goal when possible, cf. Fig.~\ref{fig:raytrace_sub2}. This design enables the PA-MPPI controller to exploit map information, allowing it to plan around obstacles and efficiently explore unknown space. For the implementation of ray tracing, we adopt the 3D Digital Differential Analyzer (DDA) algorithm~\cite{amanatides1987DDA}, which does not require a signed distance field representation of the environment. Due to the high computational cost, ray tracing is performed at the last step, $k=H-1$, of the open-loop trajectory.

%% file: equations/dynamics.tex
\begin{equation}
	\begin{aligned}
		\dot{\bm{x}} = \begin{bmatrix}
			\dot{\bm{p}}_{WB} \\
			\dot{\bm{q}}_{WB} \\
			\dot{\bm{v}}_{WB} \\
			\dot{\bm{\omega}}_{B} \\
		\end{bmatrix} = \begin{bmatrix}
		\bm{v}_{WB}\\
		\frac{1}{2} \Lambda({\omega}_B) \cdot \bm{q}_{WB}\\
		\bm{q}_{WB} \odot \bm{c}_B/m + \bm{g}_W \\
		\bm{J}^{-1}(\bm{\tau}_B - \bm{\omega}_B\times \bm{J} \cdot \bm{\omega}_B)
	\end{bmatrix}.
	\end{aligned}
 \label{eq:dynamics}
\end{equation}

%% file: sections/experiments.tex
\section{Experiments}
To evaluate the performance of PA-MPPI without considering the effect of other modules in the control loop, such as depth sensor noise or imperfect state estimation, we conduct simulated and hardware-in-the-loop (HIL) \cite{foehn2022agilicious} experiments. In the HIL setting, we use a motion capture system to acquire the ground truth state of the quadrotor, and the Flightmare simulator \cite{song2020flightmare} to render depth images in real time. The PA-MPPI controller is implemented in JAX and integrated into the Agilicious control framework \cite{foehn2022agilicious}, running on a laptop with an i7-13800H CPU with 64GB RAM and a NVIDIA A1000 laptop GPU with 6GB VRAM. The quadrotor used has a mass of \SI{0.21}{\kg} and arm length $l = \SI{19.4}{\cm}$, with propeller radius of \SI{3.81}{\cm} and a thrust to weight ratio of 6.8. 

\begin{figure}[!t]
    \centering
    \subfloat[C-wall ($w=\SI{3.0}{\meter})$]{%
        \includegraphics[height=0.125\textheight, trim=135 225 150 225, clip]{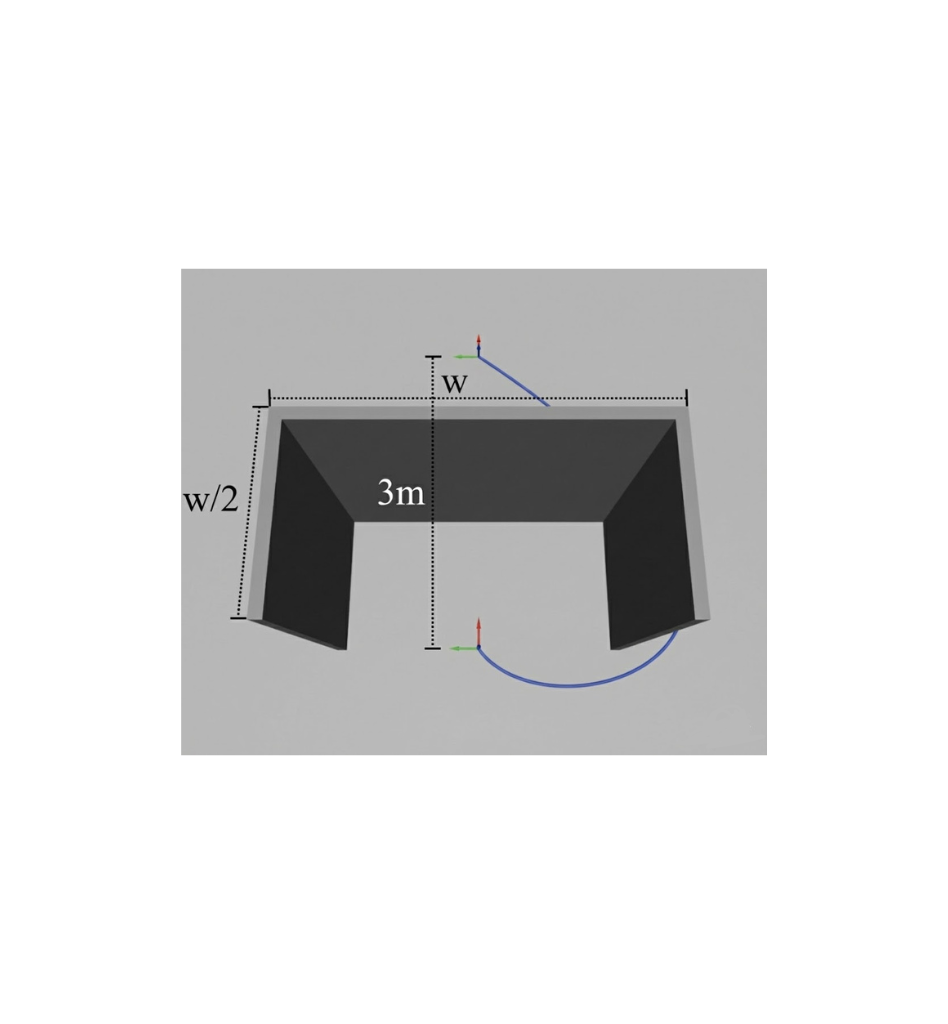}%
        \label{fig:scene_2}%
    }
    \hspace{1cm}
    \subfloat[Hole ($d=\SI{0.5}{\meter}$)]{%
        \includegraphics[height=0.125\textheight, trim=135 150 150 200, clip]{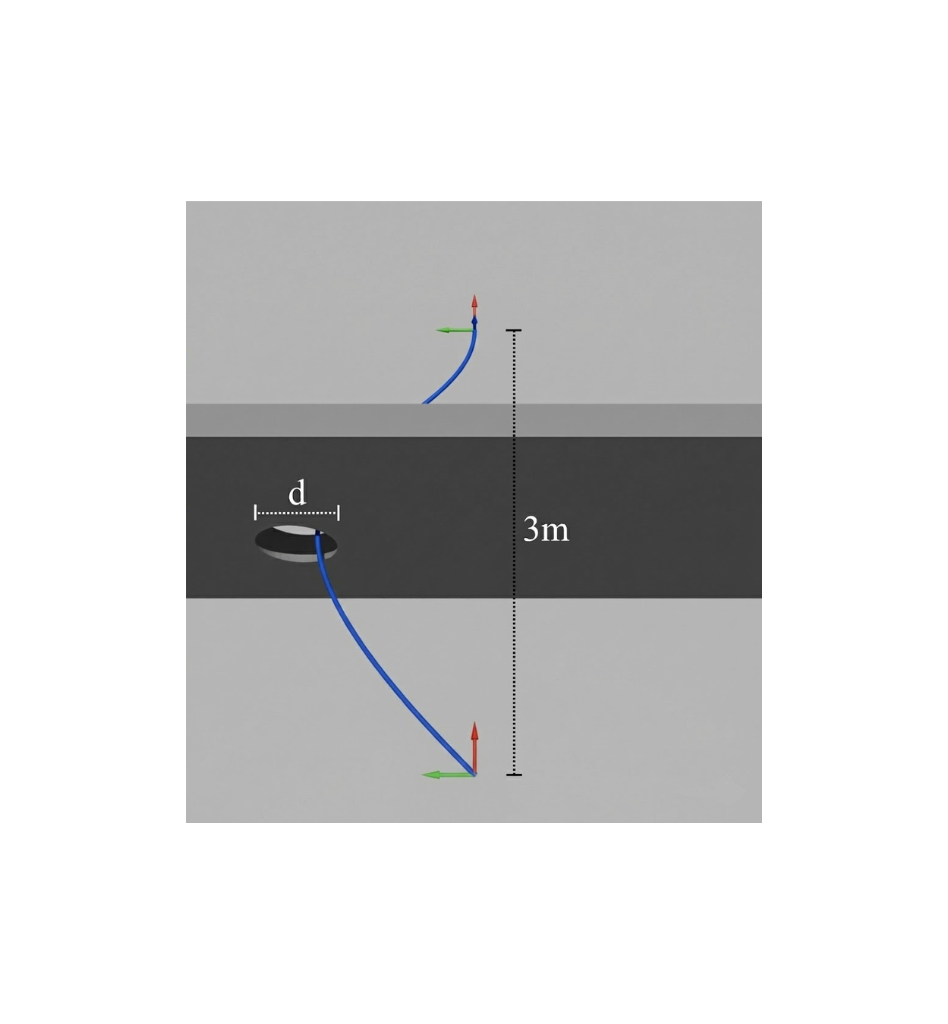}%
        \label{fig:scene_4}%
    }\hfill
    \subfloat[4-Wall ($w=\SI{1.5}{\meter}$)]{%
        \includegraphics[height=0.125\textheight, trim=0 80 0 100, clip]{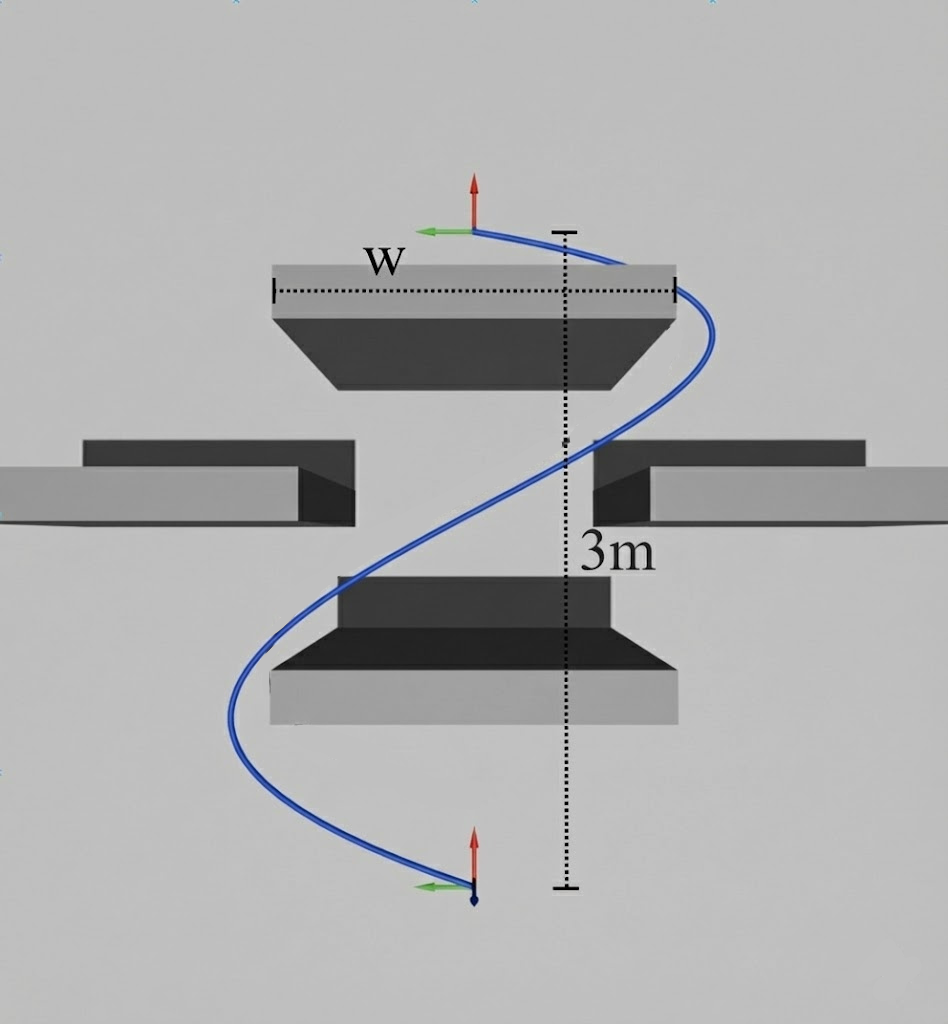}%
        \label{fig:scene_6}%
    }
    \caption{Synthetic scenes for navigation experiments. The goal pose for each task is always \SI{3}{\meter} ahead of the initial pose, with three types of obstacles in between: a C-shaped wall (a), a wall with a hole (b), and four walls (c). The most challenging setting for each scene is illustrated here, with example successful trajectories depicted in blue.}
    \label{fig:synth_scenes}
    \vspace{-0.5cm}
\end{figure}

Two sets of experiments are conducted. The first consists of navigating synthetic scenes of varying difficulty (Fig. \ref{fig:synth_scenes}) to quantitatively evaluate the performance of PA-MPPI. \reb[R7.2]{Robustness of PA-MPPI is also investigated by injecting wind disturbances during navigation.} The second consists of indoor navigation scenes from the Habitat Matterport dataset \cite{ramakrishnan2021hm3d}, using goal poses proposed by a navigation foundation model~\cite{sridhar2023nomad} to demonstrate an example usage of PA-MPPI as the action policy for a Vision-Language-Action (VLA) model in unknown environments.
A list of PA-MPPI parameters are provided in Table \ref{table:params}. As in our experiments, the weight of the terminal safe set~$c_\mathrm{safe}$ barely had an influence, we set it to zero.

\begin{table}
    % \vspace{-0.3cm}
    \begin{center}
         \caption{\textnormal{PA-MPPI parameters.}}
         \label{table:params}
         \setlength{\tabcolsep}{4pt}
         \begin{tabular}{@{}lr|lr|lr@{}}
             \toprule
             \multicolumn{2}{c}{\textbf{MPPI param.}} & \multicolumn{4}{c}{\textbf{Cost definition constants}}\\ \midrule
             \grayrow
             $N$        & \reb[R7.2] 17,500    & $c_{\text{pos}}$ & 2.5 & $c_{\text{vel}}$ & \reb[R7.2]{5.0}\\
             $H$        & 15        & $c_{\psi}$     & 1.0 & $c_{\text{unknown}}$ & \reb[R7.2]{-4.0}\\
             \grayrow
             $\lambda $ & \reb[R7.2]0.02      &  $c_{\text{collision}}$& \reb[R7.2]{15.0} & $c_{\text{occupied}}$ & \reb[R7.2]{2.0}\\
             $\Delta t_{\text{pred}}$ & \reb[R7.7]{\SI{0.1}{\second}}  & $c_{\text{PoI}}$  & 5.0 & $R$ & diag(0.01, 0.025, 0.025, 0.2)\\ 
             \grayrow
             $\Delta t_{\text{ctrl}}$ & \reb[R7.7]{\SI{0.02}{\second}}  & $c_{\text{thresh}}$& 0.5 & $R_\Delta$ & diag(0.02, 0.05, 0.05, 0.05)\\
             \bottomrule
         \end{tabular}
    \end{center}
    \vspace{-0.5cm}
\end{table}

\label{sec:syn_scene}
\setlength{\fboxsep}{0pt}
\setlength{\fboxrule}{0.7pt}

\begin{figure}[!t]
    \centering
    \subfloat[$w=\SI{0.5}{\meter}$]{%
        \includegraphics[height=0.12\textheight, trim=300 140 290 90,clip]{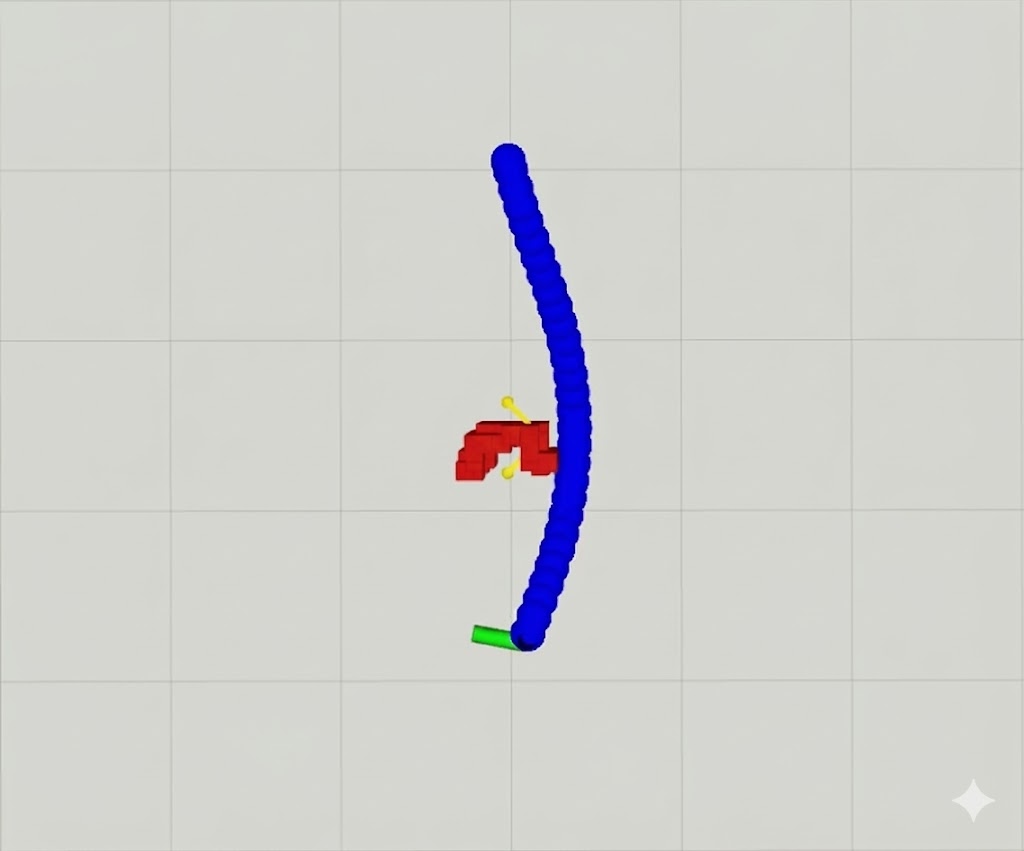}%
    } \hfill
    \subfloat[$w=\SI{2}{\meter}$]{%
        \includegraphics[height=0.12\textheight, trim=475 120 325 180,clip]{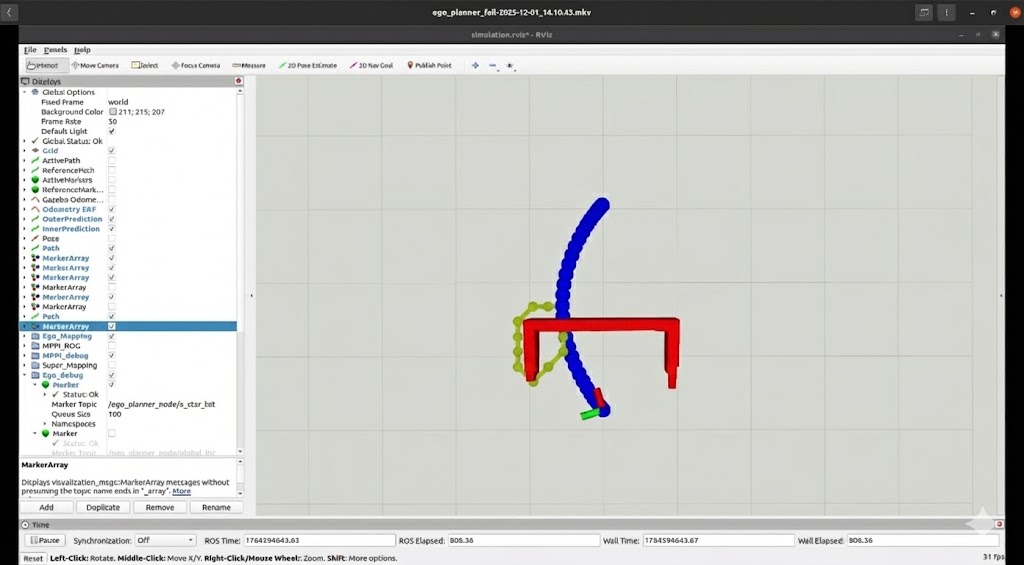}%
    } \hfill
    \subfloat[$w=\SI{2}{\meter}$]{%
        \includegraphics[height=0.12\textheight, trim=500 120 300 180,clip]{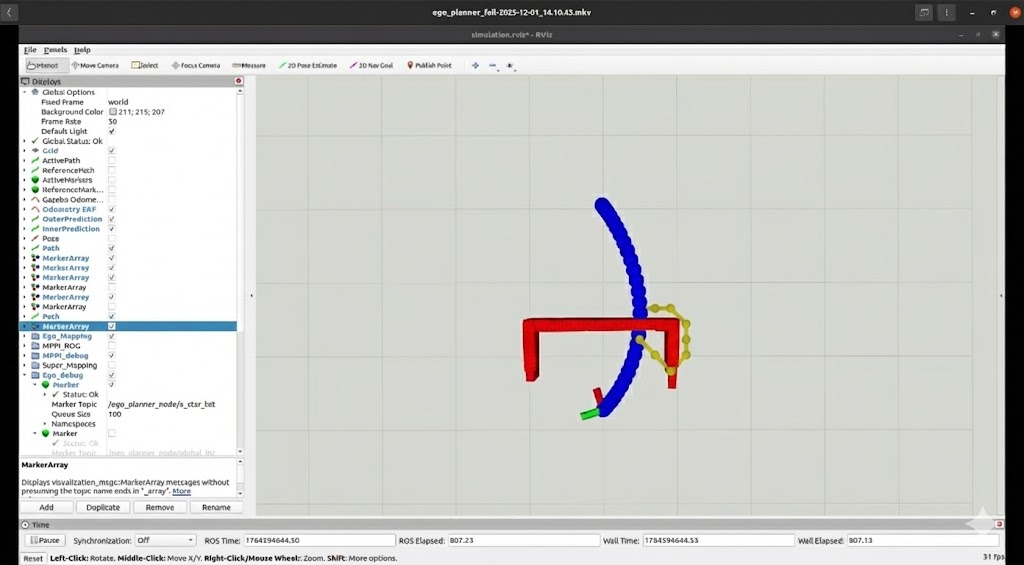}%
    }
    \caption{\reb[R5.1]{Although EGO-Planner \cite{EGO-Planner} is able to optimize the trajectory to navigate past the C-shaped wall in the easy setting (a), it struggles to find a feasible trajectory in the harder settings (b)(c). Yellow/Green: A* obstacle avoidance proposals, Blue: planned trajectories.}}
    \label{fig:ego_planner_fail}
    \vspace{-0.6cm}
\end{figure}

\textbf{Synthetic Scene Experiment.}
To quantitatively evaluate the performance of PA-MPPI, we design three scenes for navigation. The first is a C-shaped wall of varying sizes, a challenging obstacle against greedy policies. The second is a hole in a wall, similar to manholes that quadrotors must navigate through during ship inspections \cite{Dharmadhikari2023autoassess}. The third scene consists of four walls that the quadrotor must navigate past, which tests path-finding capabilities and precise control through narrow gaps of $\SI{0.5}{\meter}$. The difficulty of each scene is parametrized by the size of the obstacles $w$ (each difficulty tested 5 times) or the diameter of the holes $r$ (5 random locations, each tested 2 times), as shown in Fig.~\ref{fig:synth_scenes}. \reb[R5.1]{We validate the sufficient difficulty of the scenes by first showing that a gradient-based trajectory optimization algorithm, EGO-Planner \cite{EGO-Planner}, fails to solve the tasks at harder difficulties. As shown in Fig.~\ref{fig:ego_planner_fail}, while EGO-Planner is able to successfully navigate past a small obstacle, the optimization fails to converge when the A* obstacle avoidance front-end deviates significantly from the initial trajectories, resulting in huge feasibility costs that prevent the optimization from converging.}

\reb[R7.1, R5.1]{As EGO-Planner \cite{EGO-Planner} and trajectory tracking MPPI \cite{saska2024MPPI} cannot navigate around obstacles beyond the easiest settings, we use SUPER\cite{super}, the state-of-the-art safety-assured MAV navigation planner in unknown environments, as the baseline. %During each replanning in SUPER, two trajectories are generated: one safe backup trajectory in known free space and another in both known and unknown spaces, enabling exploration and maximizing speed. 
SUPER replans at \SI{10}{\hertz}, and a geometric controller is implemented to track the trajectories. We tune SUPER's parameter to be as fast as possible without collisions in our test scenes.}

\begin{figure}[!t]
    \centering
    \subfloat[$t=\SI{0}{\second}$]{%
        \includegraphics[height=0.1\textheight, trim=150 0 150 80, clip]{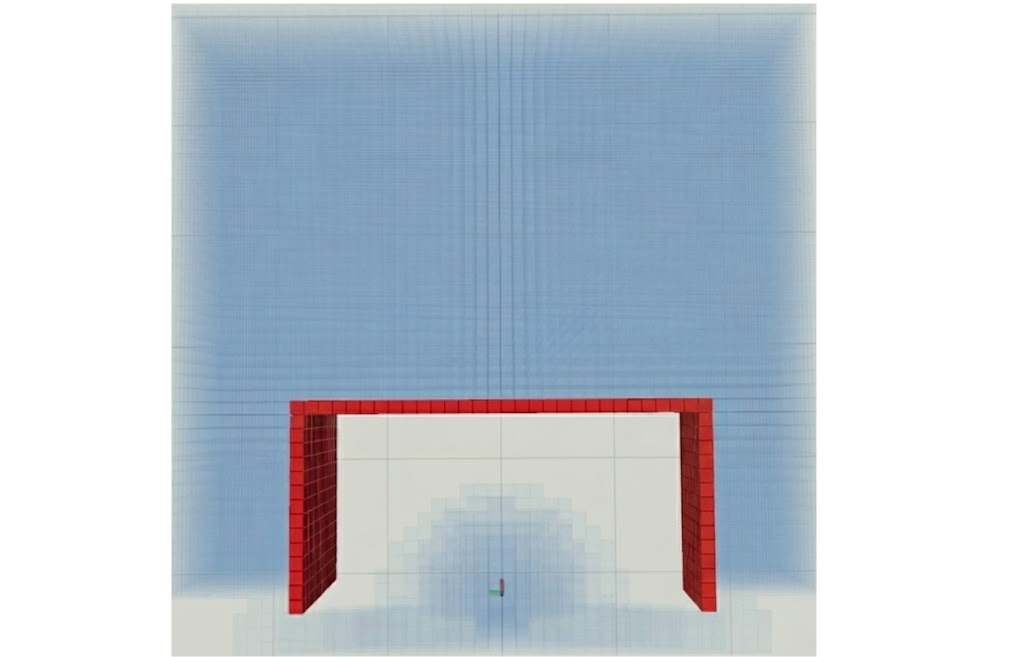}%
    }\hfill
    \subfloat[$t=\SI{3.0}{\second}$]{%
        \includegraphics[height=0.1\textheight, trim=100 0 0 300,clip]{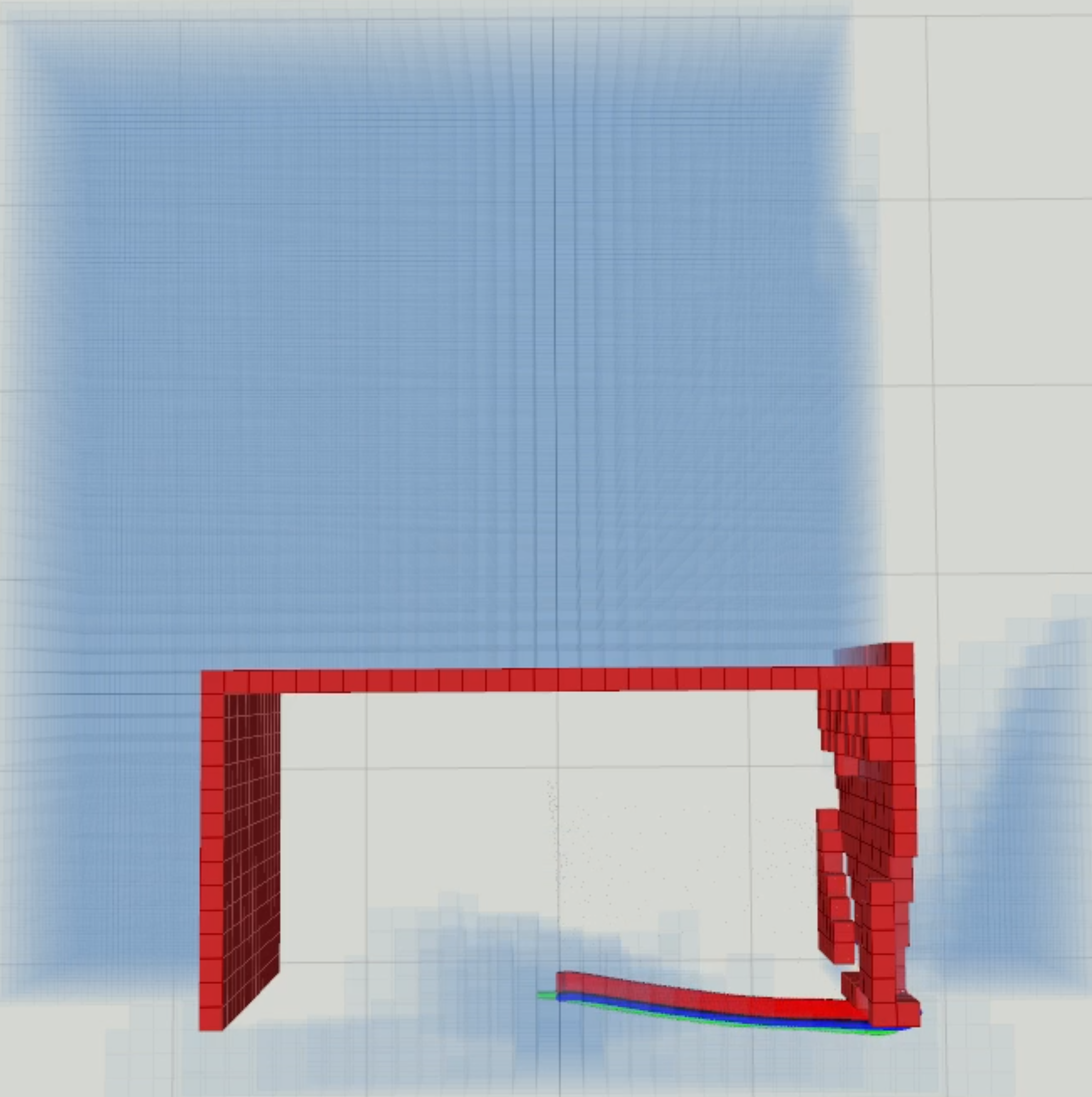}%
    }\hfill
    \subfloat[$t=\SI{5.6}{\second}$]{%
        \includegraphics[height=0.1\textheight, trim=100 0 0 180,clip]{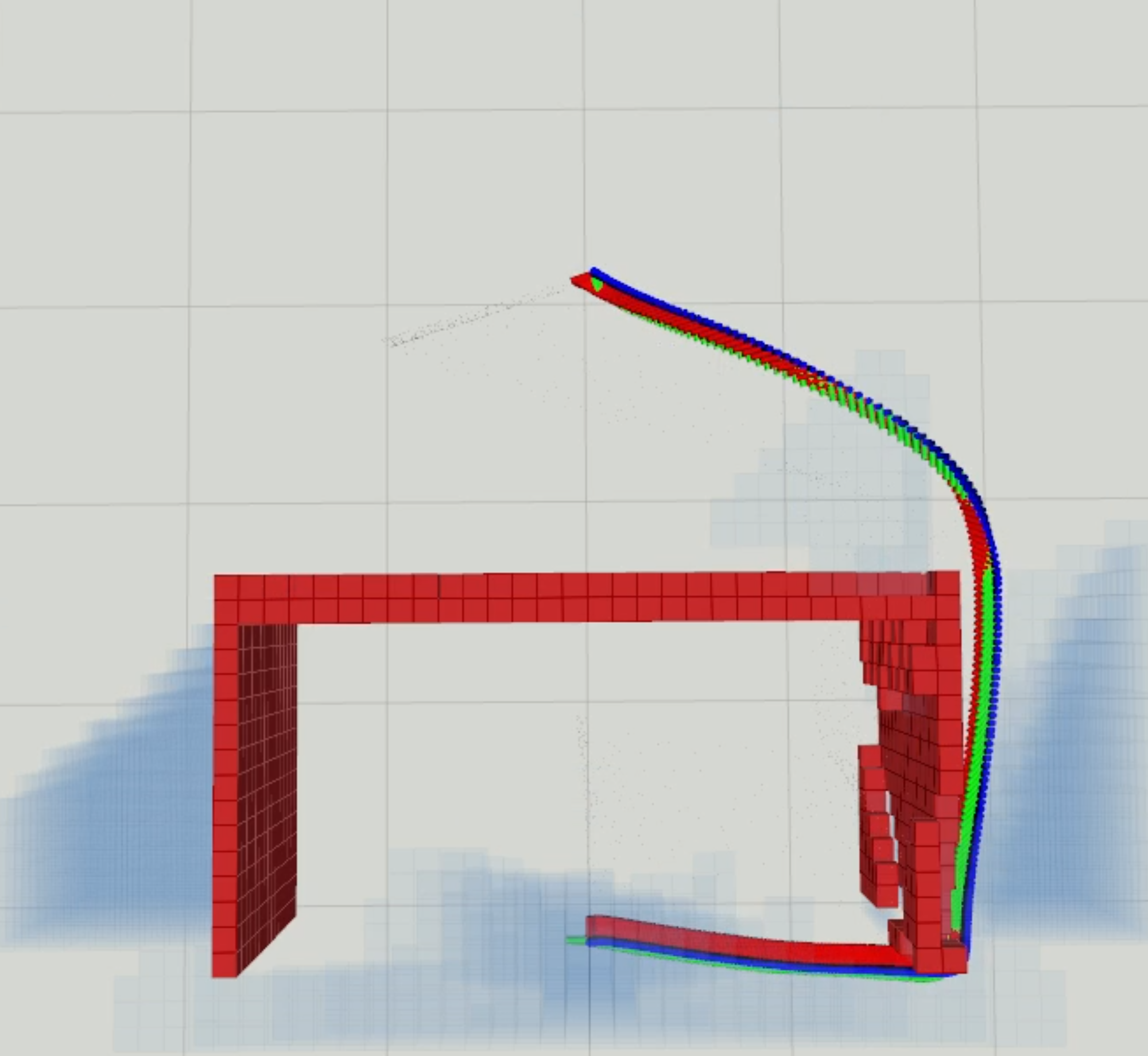}%
    }
    \caption{PA-MPPI trajectory (C-wall with $w=\SI{3.0}{\meter}$), total distance $\SI{5.7}{\meter}$.}
    \label{fig:cwall_viz}
    \vspace{-0.1cm}
\end{figure}

\begin{figure}[!t]
    \centering
    \subfloat[$t=\SI{0}{\second}$]{%
        \includegraphics[width=0.35\columnwidth, trim=0 50 0 400, clip]{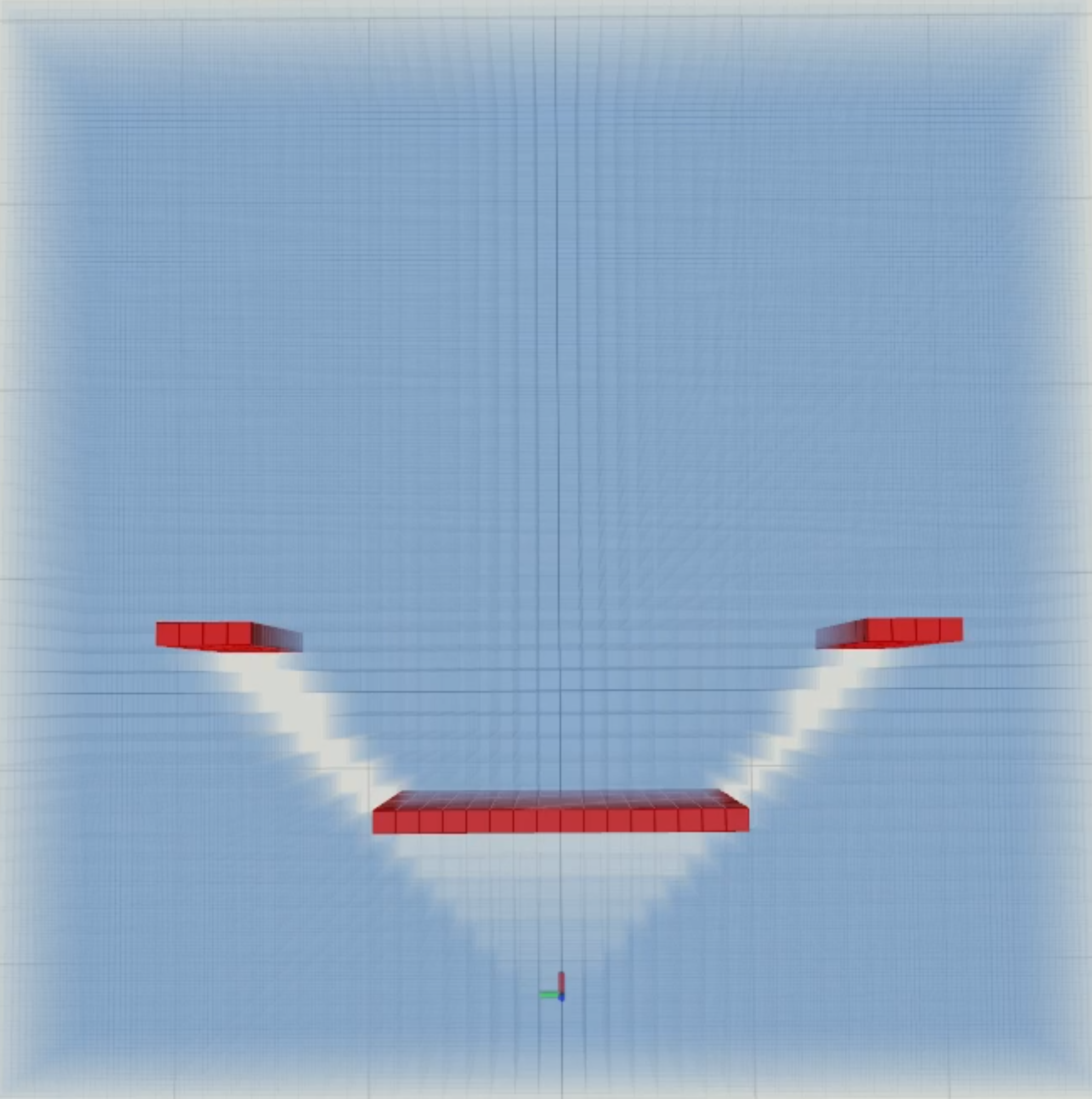}%
    }\hspace{1cm}
    \subfloat[$t=\SI{1.8}{\second}$]{%
        \includegraphics[width=0.35\columnwidth, trim=0 50 0 400, clip]{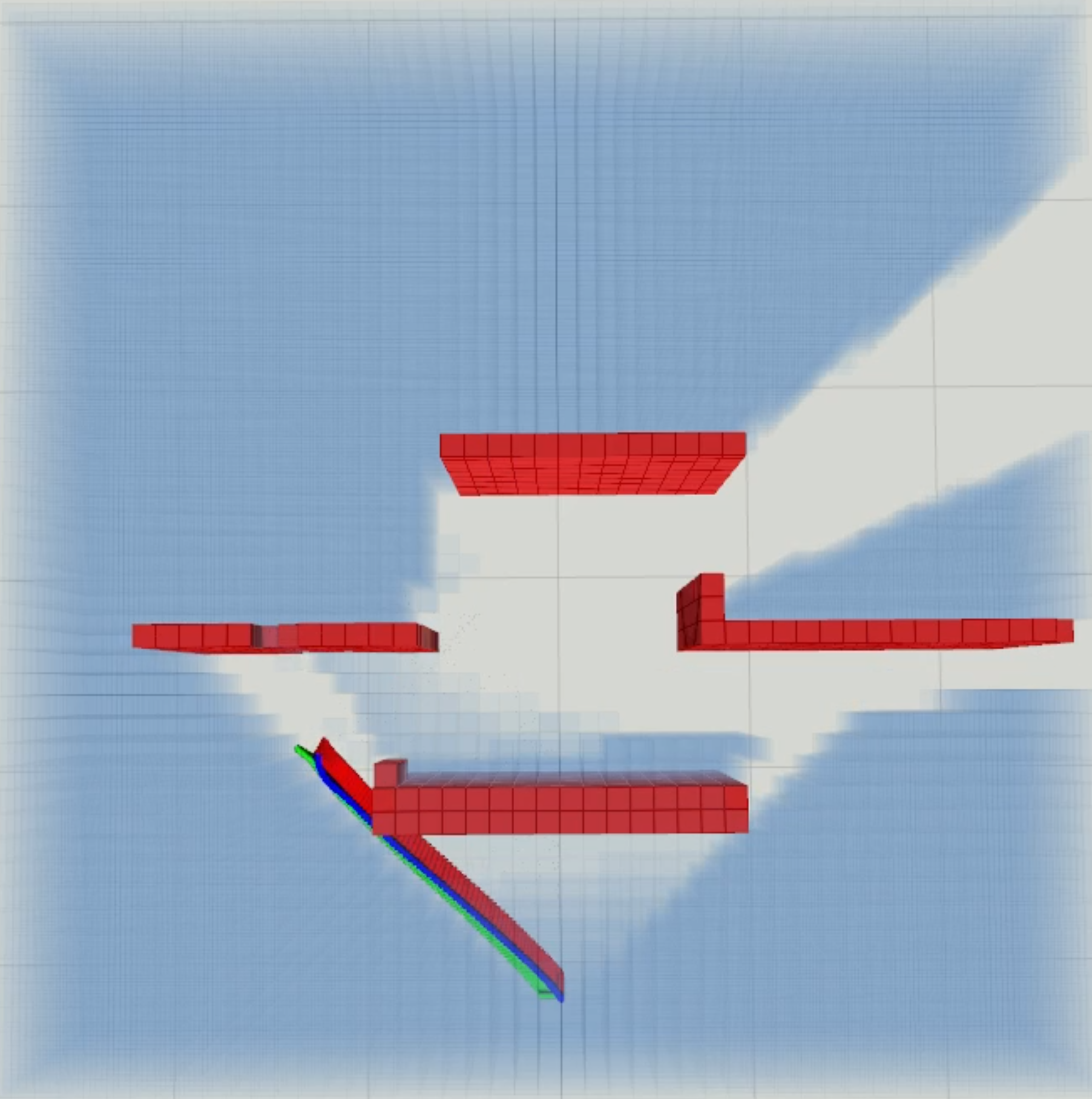}%
    }\\[1ex] % line break between rows
    \vspace{-0.4cm}
    \subfloat[$t=\SI{4.6}{\second}$]{%
        \includegraphics[width=0.35\columnwidth, trim=0 50 0 400, clip]{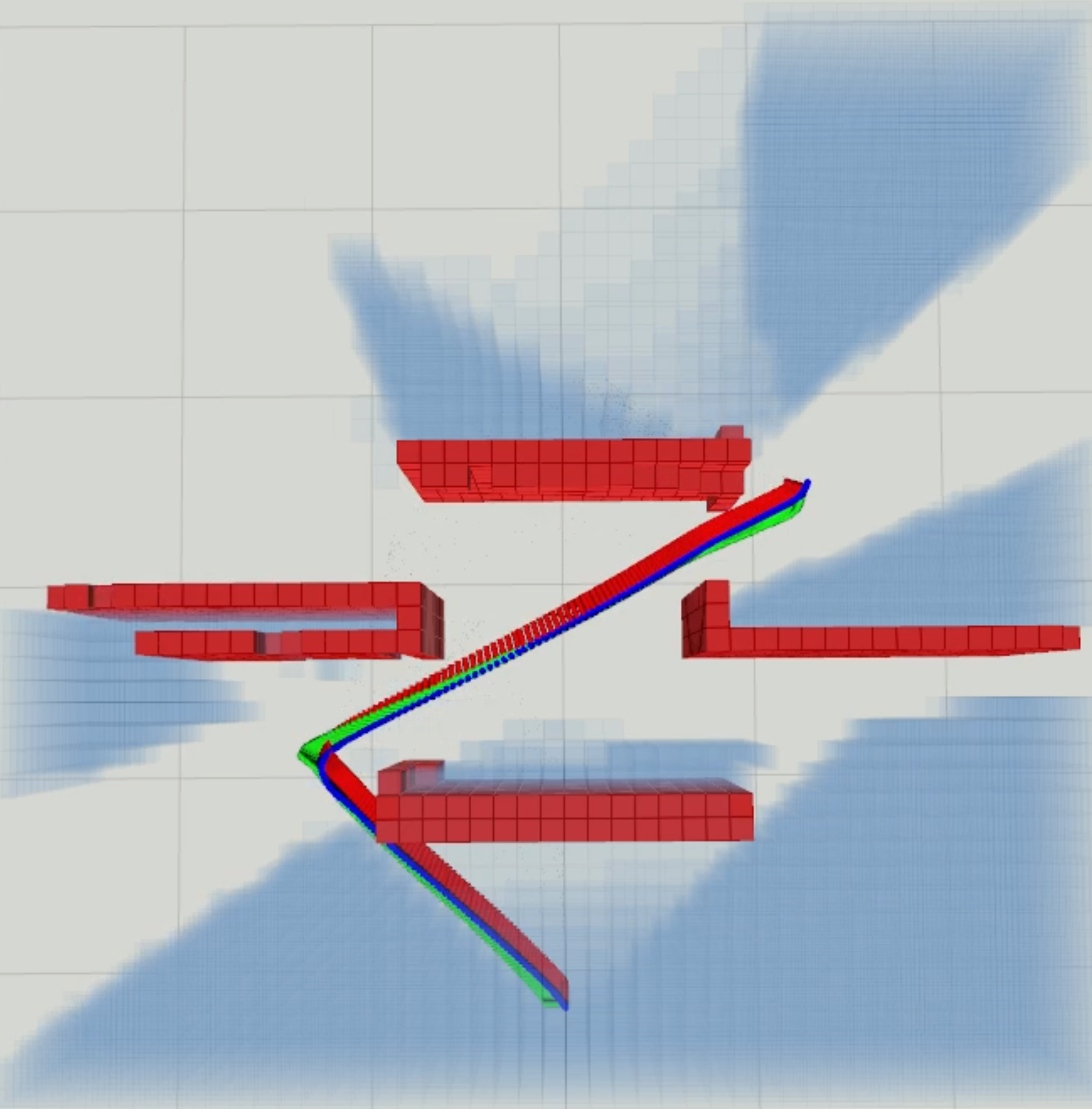}%
    }\hspace{1cm}
    \subfloat[$t=\SI{6.7}{\second}$]{%
        \includegraphics[width=0.35\columnwidth, trim=0 50 0 400, clip]{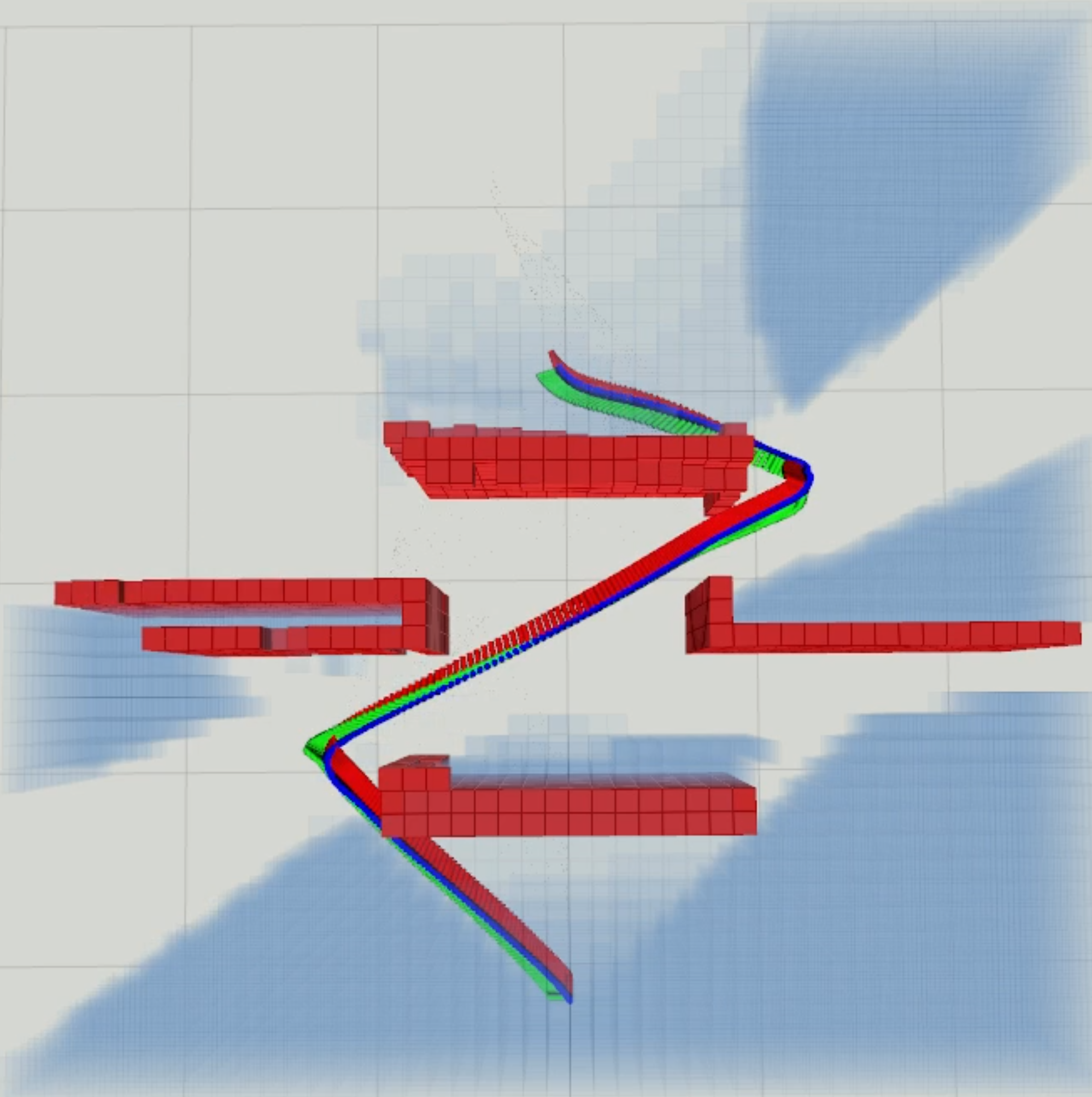}%
    }
    \caption{PA-MPPI trajectory (4-Wall with $w=\SI{1.5}{\meter}$), total distance $\SI{6.3}{\meter}$}
    \label{fig:4wall_viz}
    \vspace{-0.1cm}
\end{figure}

\begin{figure}[!t]
    \centering
    \subfloat[Duration $\SI{6.3}{\second}$, dist. \SI{6.3}{\meter}]{%
        \includegraphics[height=0.11\textheight, trim=0 0 0 370, clip]{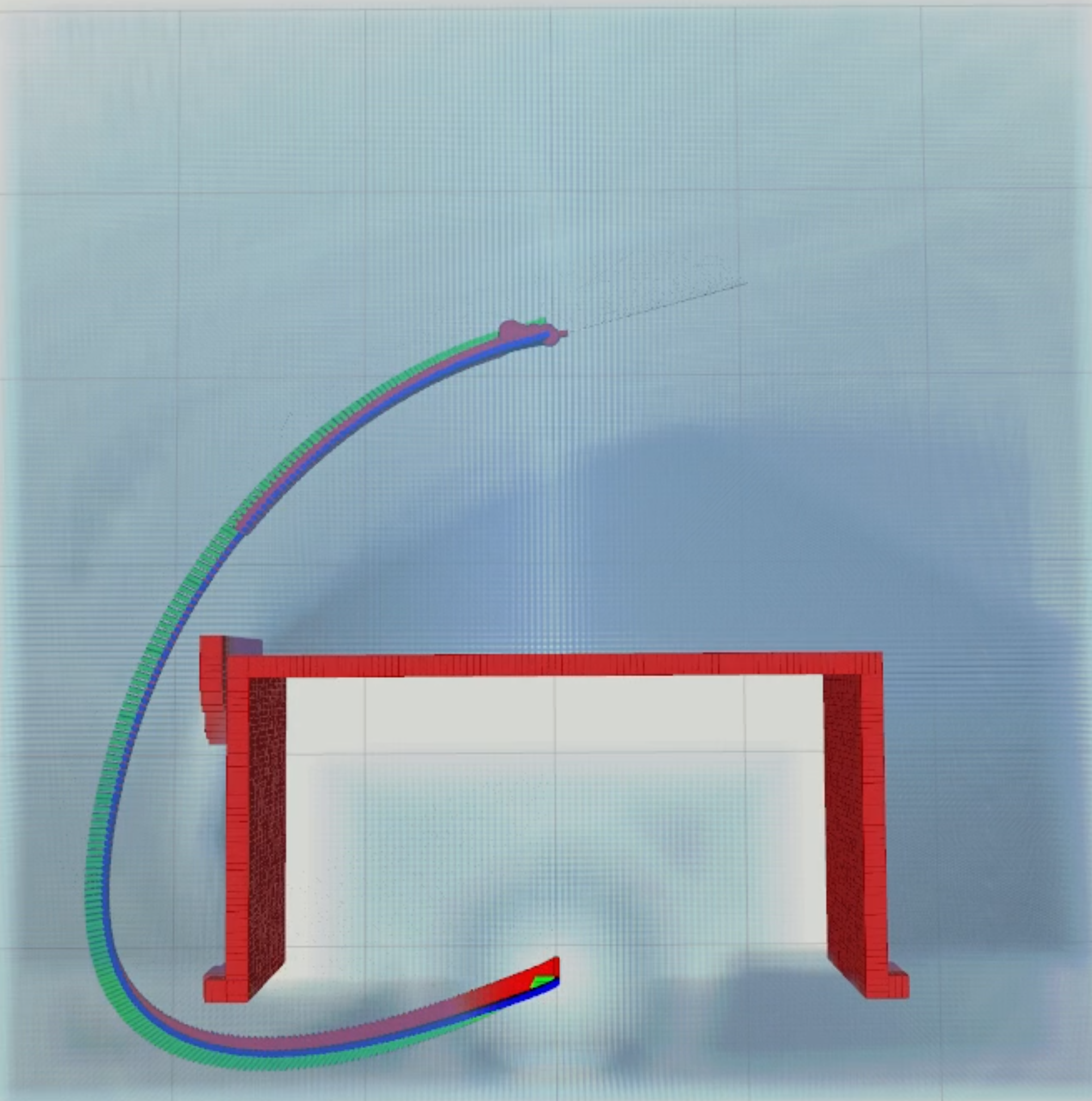}%
        \label{fig:super_cwall}
    }
    \hspace{1cm}
    \subfloat[Duration $\SI{7.6}{\second}$, dist. \SI{5.5}{\meter}]{%
        \includegraphics[height=0.11\textheight, trim=0 0 0 370,clip]{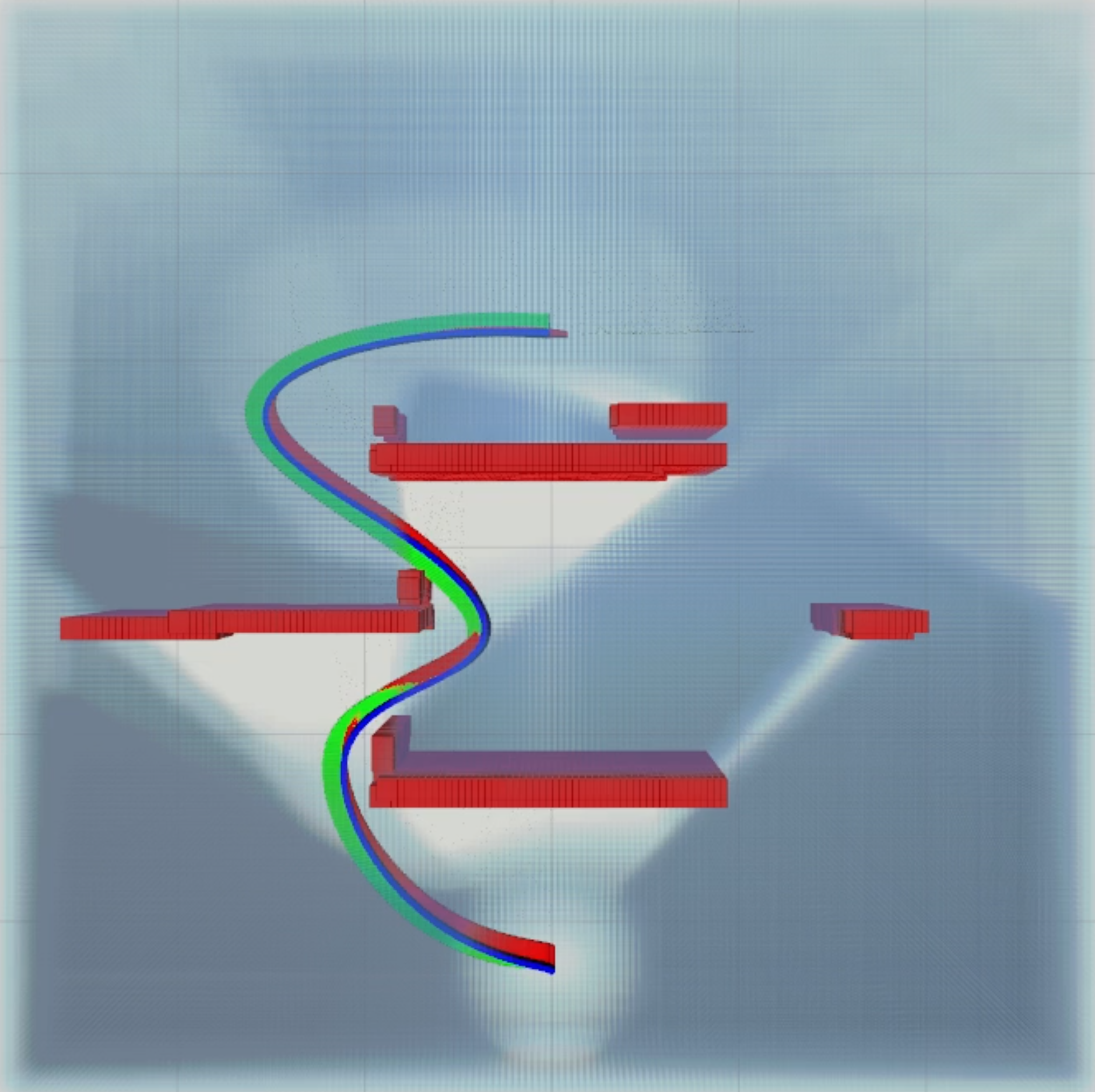}%
        \label{fig:super_wall4}
    }
    \caption{SUPER trajectory samples in the C-wall and the 4-wall scene}
    \label{fig:super_viz}
    \vspace{-0.1cm}
\end{figure}

\reb[R7.1, R5.1]{The experiment results are summarized in Table \ref{table:c_wall_res}, \ref{table:hole_res}, \ref{table:4_wall_res}. Guided by the perception cost, PA-MPPI moves to positions that help navigate around obstacles and map the environment towards the goal, successfully completing the tasks, as visualized in Fig.~\ref{fig:cwall_viz} and~\ref{fig:4wall_viz}. While both SUPER and PA-MPPI are able to complete the tasks with 100\% success rate, PA-MPPI consistently performs better in terms of total time and velocity of the trajectories, while having trajectory distances on par with SUPER. For a qualitative comparison, we visualize two trajectories from SUPER in Fig.~\ref{fig:super_viz}. For the C-wall scene, although SUPER's A* front-end proposed trajectories close to the obstacle wall, the subsequent trajectory optimization aggressively optimized for dynamic feasibility by penalizing jerk, body rate, and collective thrust, resulting in a smoother but longer trajectory, as shown in Fig.~\ref{fig:super_cwall}. In the 4-wall scene, the trajectory adheres to the shortest path in distance suggested by the A* front end. However, compared to PA-MPPI's trajectory, the final trajectory requires significantly more rotational control effort, as A* is unaware of the quadrotor dynamics. 
% As SUPER is guided by the A* front-end trajectory computed on an inflated occupancy map, and subsequently optimized to track it while satisfying feasibility, this decoupling introduces potential suboptimality in the final trajectory: the A* proposal does not consider the quadrotor dynamics or time optimality, and the resulting trajectory that adheres to it, while feasible, is potentially suboptimal, resulting in lower tracking velocities and in some cases, overall longer trajectories. 
On the other hand, PA-MPPI inherently samples dynamically feasible trajectories, allowing it to prioritize trajectory quality in the sampling-based optimization. As a result, PA-MPPI's trajectories are faster and significantly more energy-efficient.}

\begin{table}[!t]
    \centering
    \caption{\textnormal{\reb[R7.1, R5.1]{C-Shaped Wall Experiment Results.}}}
    \label{table:c_wall_res}
    
    % 1. CHANGE FONT SIZE HERE
    % Options: \small, \footnotesize (standard), \scriptsize (very small), \tiny (hard to read)
    \scriptsize 
    
    % 2. TIGHTEN ROW SPACING (Optional, default is 1.0)
    \renewcommand{\arraystretch}{0.9} 
    
    \begin{tabular*}{\columnwidth}{@{\extracolsep{\fill}}cc|cccc@{}}
        \toprule
         & \makecell{\textbf{Obs.} \\ \textbf{Size} \\ $w$} 
         & \makecell{\textbf{Dist.} \\ (\si{\meter})} 
         & \makecell{\textbf{Time} \\ (\si{\second})} 
         & \makecell{\textbf{Avg.} \\ \textbf{Vel.} \\ (\si{\meter\per\second})}
         & \makecell{\textbf{Energy} \\ \textbf{Consumption} \\ (\si{\joule})} \\ 
        \midrule
        
        \multirow{3}{*}{\makecell{SUPER \\ (sim)}} % & \SI{0.5}{\meter} & & & & & \\
         & \SI{1.0}{\meter} & $3.57\pm0.15$ & $4.54\pm0.51$ & $0.79\pm0.05$ & $217.1\pm25.3$\\
         & \SI{2.0}{\meter} & $4.54 \pm 0.12$ & $4.72\pm0.28$ & $0.96\pm0.03$ & $225.0\pm13.2$\\
         & \SI{3.0}{\meter} & $6.48\pm0.92$ & $6.00\pm0.84$ & $1.08\pm0.03$ & $286.8\pm42.0$\\
        \midrule
        
        \multirow{3}{*}{\makecell{PA-MPPI \\ (sim)}} % & \SI{0.5}{\meter} & & & & & \\
         & \SI{1.0}{\meter} & $3.56\pm0.04$ & $2.91\pm0.04$ & $1.22\pm0.03$ & $142.0\pm2.0$\\
         & \SI{2.0}{\meter} & $4.56\pm0.07$ & $4.20\pm0.07$ & $1.09\pm0.01$ & $202.6\pm3.5$ \\
         & \SI{3.0}{\meter} & $5.87\pm0.07$ & $5.14\pm0.22$ & $1.14\pm0.06$ & $249.5\pm9.5$\\
        \midrule
        
        \multirow{3}{*}{\makecell{PA-MPPI \\ (real)}} % & \SI{0.5}{\meter} & & & & & \\
         & \SI{1.0}{\meter} & $3.61\pm0.13$ & $2.84\pm0.05$ & $1.27\pm0.04$ & \textemdash\\
         & \SI{2.0}{\meter} & $4.59\pm0.10$ & $4.09\pm0.09$ & $1.12\pm0.02$ & \textemdash\\
         & \SI{3.0}{\meter} & $5.88\pm0.11$ & $5.13\pm0.18$ & $1.15\pm0.06$ & \textemdash\\
        \bottomrule
    \end{tabular*}
\end{table}

\begin{table}[!t]
    \centering
    \caption{\textnormal{\reb[R7.1, 5.1]{Hole in the Wall Experiment Results.}}}
    \label{table:hole_res}
    
    % 1. CHANGE FONT SIZE HERE
    % Options: \small, \footnotesize (standard), \scriptsize (very small), \tiny (hard to read)
    \scriptsize 
    
    % 2. TIGHTEN ROW SPACING (Optional, default is 1.0)
    \renewcommand{\arraystretch}{0.9} 
    
    \begin{tabular*}{\columnwidth}{@{\extracolsep{\fill}}cc|cccc@{}}
        \toprule
         & \makecell{\textbf{Hole} \\ \textbf{Diameter} \\ $d$} 
         & \makecell{\textbf{Dist.} \\ (\si{\meter})} 
         & \makecell{\textbf{Time} \\ (\si{\second})} 
         & \makecell{\textbf{Avg.} \\ \textbf{Vel.} \\ (\si{\meter\per\second})}
         & \makecell{\textbf{Energy} \\ \textbf{Consumption} \\ (\si{\joule})} \\ 
        \midrule
        
        \multirow{2}{*}{\makecell{SUPER \\ (sim)}} % & \SI{0.5}{\meter} & & & & & \\
         & \SI{0.5}{\meter} & $3.77\pm0.41$ & $4.36\pm0.44$ & $0.87\pm0.05$ & $209.0\pm21.8$\\
         & \SI{1.0}{\meter} & $3.83\pm0.79$ & $4.26\pm0.56$ & $0.89\pm0.08$ & $203.4\pm27.3$\\
        \midrule
        
        \multirow{2}{*}{\makecell{PA-MPPI \\ (sim)}} % & \SI{0.5}{\meter} & & & & & \\
         & \SI{0.5}{\meter} & $3.66\pm0.40$ & $3.23\pm0.42$ & $1.14\pm0.07$ & $155.7\pm20.3$\\
         & \SI{1.0}{\meter} & $3.44\pm0.38$ & $2.79\pm0.35$ & $1.24\pm0.08$ & $135.8\pm17.1$\\
        \midrule
        
        \multirow{2}{*}{\makecell{PA-MPPI \\ (real)}} % & \SI{0.5}{\meter} & & & & & \\
         & \SI{0.5}{\meter} & $3.71\pm0.45$ & $3.22\pm0.65$ & $1.17\pm0.11$ & \textemdash\\
         & \SI{1.0}{\meter} & $3.52\pm0.37$ & $2.67\pm0.30$ & $1.32\pm0.10$ & \textemdash\\
        \bottomrule
    \end{tabular*}
\end{table}

\begin{table}[!t]
    \centering
    \caption{\textnormal{\reb[R7.1, R5.1]{4-Wall Experiment Results.}}}
    \label{table:4_wall_res}
    
    % 1. CHANGE FONT SIZE HERE
    % Options: \small, \footnotesize (standard), \scriptsize (very small), \tiny (hard to read)
    \scriptsize 
    
    % 2. TIGHTEN ROW SPACING (Optional, default is 1.0)
    \renewcommand{\arraystretch}{0.9} 
    
    \begin{tabular*}{\columnwidth}{@{\extracolsep{\fill}}cc|cccc@{}}
        \toprule
         & \makecell{\textbf{Obs.} \\ \textbf{Size} \\ $w$} 
         & \makecell{\textbf{Dist.} \\ (\si{\meter})} 
         & \makecell{\textbf{Time} \\ (\si{\second})} 
         & \makecell{\textbf{Avg.} \\ \textbf{Vel.} \\ (\si{\meter\per\second})}
         & \makecell{\textbf{Energy} \\ \textbf{Consumption} \\ (\si{\joule})} \\ 

        \midrule
        
        \multirow{3}{*}{\makecell{SUPER \\ (sim)}} % & \SI{0.5}{\meter} & & & & & \\
         & \SI{0.5}{\meter} & $3.42\pm0.14$ & $4.30\pm0.14$ & $0.79\pm0.02$ & $204.7\pm6.1$\\
         & \SI{1.0}{\meter} & $5.22\pm0.60$ & $6.53\pm0.34$ & $0.80\pm0.08$ & $313.8\pm17.0$\\
         & \SI{1.5}{\meter} & $5.47\pm1.25$ & $7.57\pm0.99$ & $0.72\pm0.07$ & $364.3\pm50.0$\\
        \midrule
        
        \multirow{3}{*}{\makecell{PA-MPPI \\ (sim)}} % & \SI{0.5}{\meter} & & & & & \\
         & \SI{0.5}{\meter} & $3.78\pm0.20$ & $3.12\pm0.08$ & $1.19\pm0.05$ & $155.1\pm4.1$\\
         & \SI{1.0}{\meter} & $3.97\pm0.12$ & $4.52\pm0.15$ & $0.88\pm0.03$ & $216.7\pm7.0$\\
         & \SI{1.5}{\meter} & $6.03\pm0.33$ & $5.78\pm0.23$ & $1.05\pm0.07$ & $281.7\pm11.2$\\
        \midrule
        
        \multirow{3}{*}{\makecell{PA-MPPI \\ (real)}} % & \SI{0.5}{\meter} & & & & & \\
         & \SI{0.5}{\meter} & $3.69\pm0.09$ & $2.94\pm0.13$ & $1.26\pm0.05$ & \textemdash\\
         & \SI{1.0}{\meter} & $3.95\pm0.07$ & $4.03\pm0.10$ & $0.98\pm0.04$ & \textemdash\\
         & \SI{1.5}{\meter} & $5.89\pm0.32$ & $6.75\pm0.75$ & $0.88\pm0.10$ & \textemdash\\
        \bottomrule
    \end{tabular*}
\end{table}

\begin{figure}[!t]
    \centering
    \includegraphics[width=0.85\columnwidth, trim=0 0 0 0, clip]{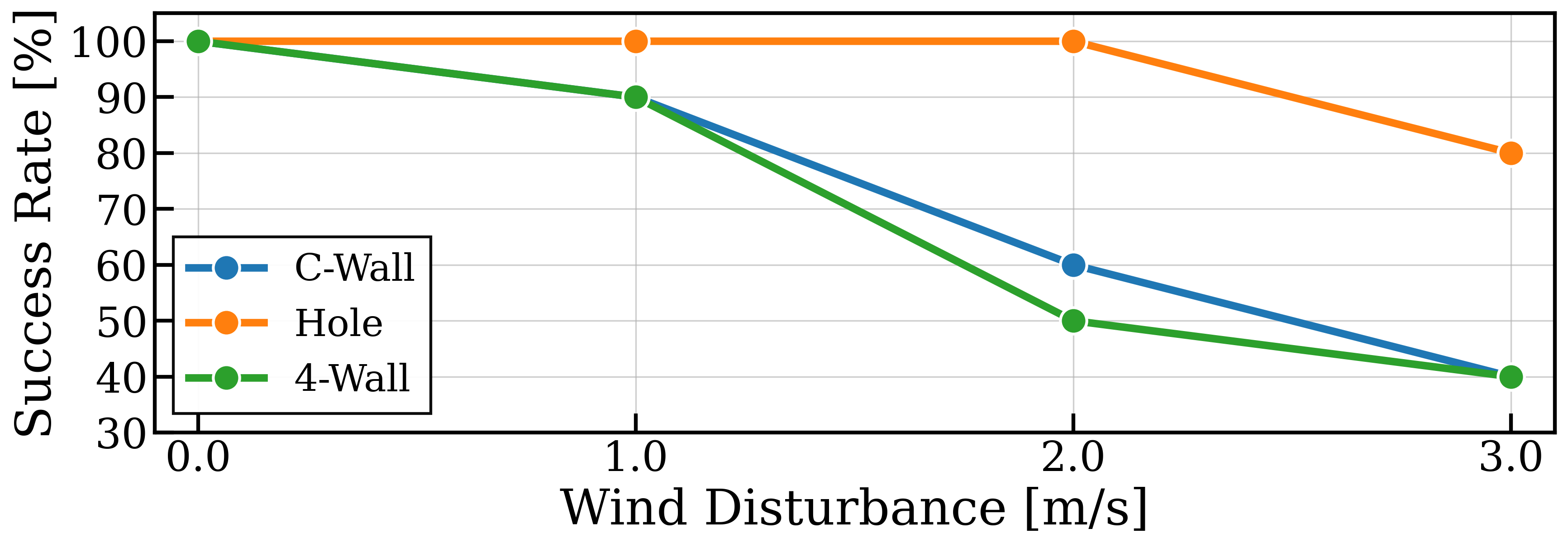}%
    \caption{\reb[R7.2]{Success rate of PA-MPPI under wind disturbance}}
    \label{fig:robustness}
    \vspace*{-0.4cm}
\end{figure}

\reb[R7.2]{
\textbf{Robustness Study.} To investigate the robustness of PA-MPPI, we conduct simulated test in the most challenging setting of the three tasks while the quadrotor is subjected to external wind disturbances. The wind directions are randomly sampled in the $xy$-plane for each episode, with wind magnitudes of $[1.0\si{\meter\per\second}, 2.0\si{\meter\per\second}, 3.0\si{\meter\per\second}]$ with 10 test episodes each. The results are shown in Fig. \ref{fig:robustness}. With low external wind speed, PA-MPPI retains a high success rate, while at larger wind speeds, the controller cannot compensate for the disturbance, resulting in more episodes failing by colliding into the obstacles or being blown into unknown areas, resulting in early terminations.
}

\begin{figure}[!t]
    \centering
    \subfloat[NoMaD proposed traj.]{%
        \includegraphics[height=0.135\textheight, trim=0 0 0 0, clip]{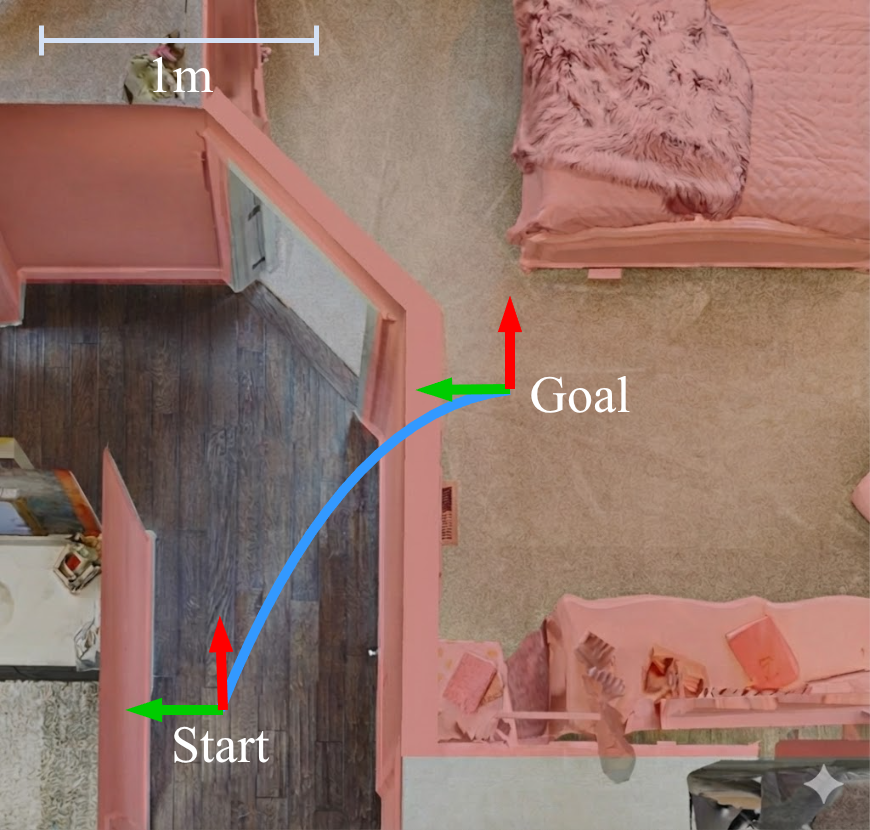}%
        \label{fig:scene1_nomad}
    } \hspace{0.2cm}
    % \vspace{-0.3cm}
    % \subfloat[Tracking MPPI]{%
    %     \includegraphics[width=0.35\columnwidth, trim=250 400 200 0,clip]{figures/nomad/scene1_baseline.png}%
    % } \hspace{1cm}
    \subfloat[PA-MPPI trajectory]{%
        \includegraphics[height=0.135\textheight, trim=200 25 0 225,clip]{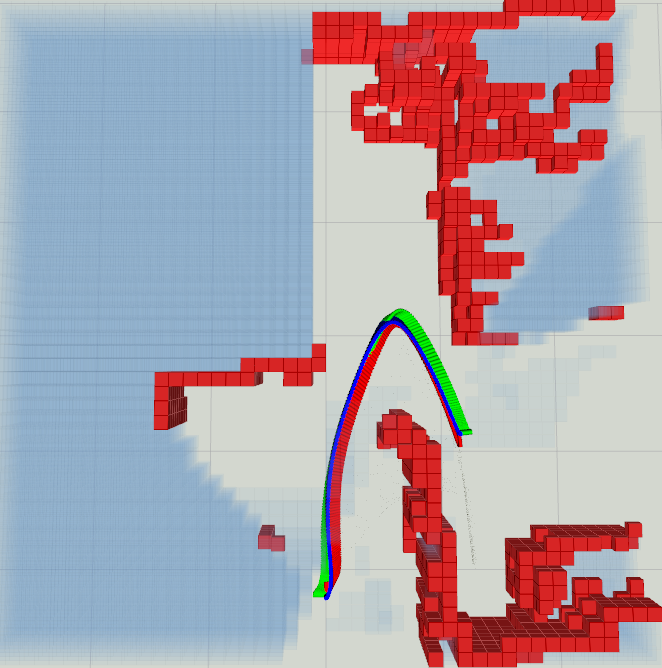}%
        \label{fig:scene1_mppi}
    } \hfill
    
    \subfloat[NoMaD proposed traj.]{%
        \includegraphics[height=0.11\textheight, trim=10 0 0 0, clip]{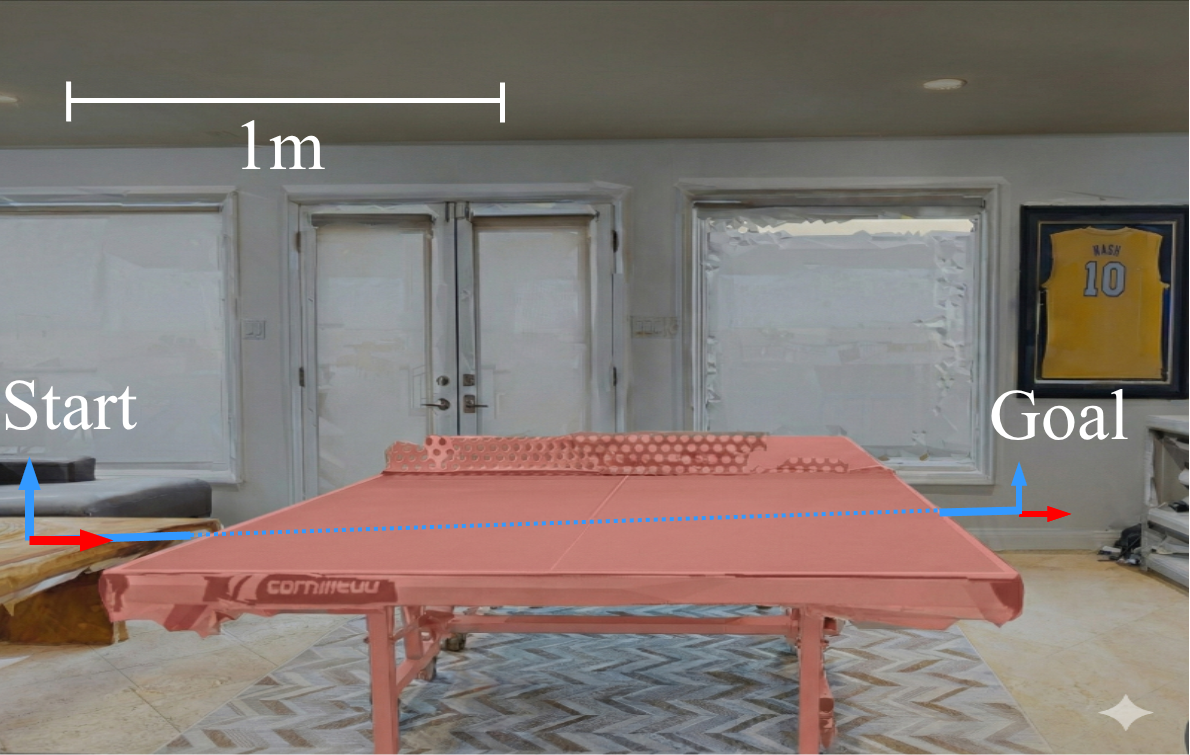}%
        \label{fig:scene2_nomad}
    } \hfill
    % \vspace{-0.3cm}
    % \subfloat[Tracking MPPI]{%
    %     \includegraphics[width=0.45\columnwidth, trim=650 600 750 400,clip]{figures/nomad/pingpoing_baseline.png}%
    % }\hfill
    \subfloat[PA-MPPI trajectory]{%
        \includegraphics[height=0.11\textheight, trim=170 70 190 0,clip]{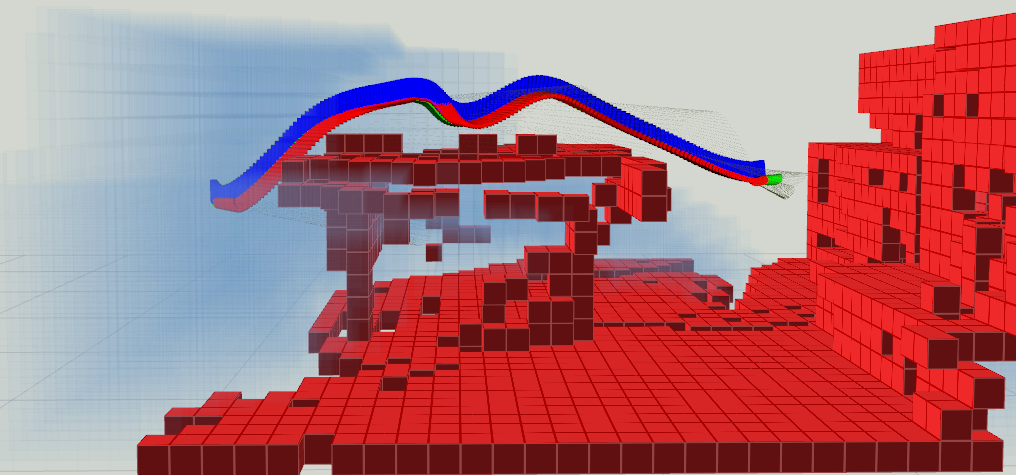}%
        \label{fig:scene2_mppi}
    }
    \caption{\reb[R7.4]{Two Habitat scenes, with obstacles (walls, furniture, etc) overlaid in red and NoMaD proposed trajectory in blue (a)(c), and visualization of the PA-MPPI trajectory in the occupancy map (b)(d).}}
    \label{fig:habitat_scenes}
    \vspace*{-0.2cm}
\end{figure}

\textbf{PA-MPPI combined with Navigation Foundation Model.}
We validate the real-world applicability of the PA-MPPI controller by using it as the action policy for a navigation foundation model, NoMaD \cite{sridhar2023nomad}, which proposes trajectory waypoints. Since the inputs to NoMaD are only monocular RGB images, the proposed waypoints are not guaranteed to be feasible due to scale ambiguity. We use two scenes from the Habitat Matterport dataset \cite{ramakrishnan2021hm3d}. In the first scene, visualized in Fig. \ref{fig:scene1_nomad}, the NoMaD proposed trajectory attempts to enter the room but misses the door. PA-MPPI was able to explore and map unknown regions in this scene, navigate through the door, and ultimately reach the goal position, as shown in Fig. \ref{fig:scene1_mppi}. In the second scene, visualized in Fig. \ref{fig:scene2_nomad}, the NoMaD trajectory starts close to a ping-pong table and goes directly through it. PA-MPPI successfully avoids the obstacle. Both scenarios demonstrate successful reference-free navigation in cluttered environments using PA-MPPI, making it a suitable action policy for navigation foundation models.

\vspace{-0.2cm}
\section{Discussion \& Future Work} 
Due to the limited FoV of the depth sensor, a good initial observation is crucial for successful navigation around large obstacles. For example, in the experiment shown in Fig. \ref{fig:cwall_viz}, the initial observation at $t=\SI{0}{\second}$ is initialized by turning the quadrotor $\pm 90^\circ$ to observe free space outside the convex hull of the C-shaped wall. However, this issue can be solved by replacing the depth camera with LIDAR, which only requires modification to the mapping module. Additonally, we only considered navigation tasks within a fixed 3D boundary. Future work may take advantage of the local map sliding feature of the ROG-Map to implement larger scale navigation. Finally, PA-MPPI’s planning horizon is constrained by available computation, which fundamentally limits the complexity of navigation scenes it can handle. This may be mitigated by using an appropriate terminal value function on the last position of the horizon. We leave this for future work to investigate.

%% file: sections/conclusion.tex
\section{Conclusion}

This work presents a perception-aware MPPI controller that integrates a novel perception-driven cost to enable reference-free quadrotor navigation in partially known, cluttered environments with MPPI. By exploiting the current map, the perception term steers trajectories toward informative frontiers to explore the unknown regions and advance towards the goal. Simulated and real-world experiments demonstrate that its performance is comparable to that of the state-of-the-art navigation planner. We further demonstrate its potential as an action policy for foundation models to navigate challenging environments. Future work will focus on extending PA-MPPI to longer-horizon navigation and conducting an in-depth study of its path-planning limits and methods to mitigate them.
% \vspace{-0.2cm}